\documentclass[10pt,twocolumn,letterpaper]{article}
\usepackage[pagenumbers]{cvpr}
\usepackage[T1]{fontenc}
\usepackage{graphicx,booktabs,tabularx,array,amsmath,amssymb}
\usepackage{enumitem}
\usepackage{microtype}
\usepackage[pagebackref,breaklinks,colorlinks,allcolors=blue]{hyperref}
\setlist{nosep,leftmargin=*}
\title{Faithful Where You Look: Anchor-Based Evaluation Hides Text Fabrication in Image-Conditioned Video Generation}
\author{Yujia Wang\\New York University Shanghai\\{\tt\small yw7997@nyu.edu}}
\begin{document}
\maketitle
\begin{abstract}
Image-conditioned video models are used to show real places, so whether their output stays faithful to the place they were shown is a practical question. The text benchmarks we found that check fidelity against a reference do so at an \emph{anchor} --- an annotated region, typically the sign the task is about. We show that this choice hides most of the failure. We conditioned four models from three vendors on 48 street-level references of Tokyo storefronts and measured text fidelity at the annotated sign and in the periphery, the other text the reference covers. Median anchor text fidelity was 0.90--0.94 for the four models, against 0.996 for a no-model control. In the periphery, \textbf{36--54\% of scored text regions were unsupported by the reference}, against 1.6\% for the control, although every prompt forbade new readable text. Two blinded raters agreed on which regions were fabricated, with $\kappa$ of 0.83 on the development pack, 0.91 on held-out Wan, and 0.96 on held-out Veo. In the preregistered held-out rounds the unsupported rates were 42.0\% for Wan and 55.5\% for Veo. VBench-I2V's reference-based consistency scores showed no evidence of tracking fabrication in our sample. Most fabricated text is legible-looking text where the reference shows none or cannot be read, which is why our initial string-matching measure missed it. A preregistered Latin-script replication in London missed its criterion in the first round and met it in an independent second round. We release code, annotations, per-video scores, ratings and the benchmark preregistrations. Code and data: \url{https://github.com/geoywang/refanchor-bench}.
\end{abstract}
\section{Introduction}

\begin{figure*}[t]
\centering
\includegraphics[width=0.96\textwidth]{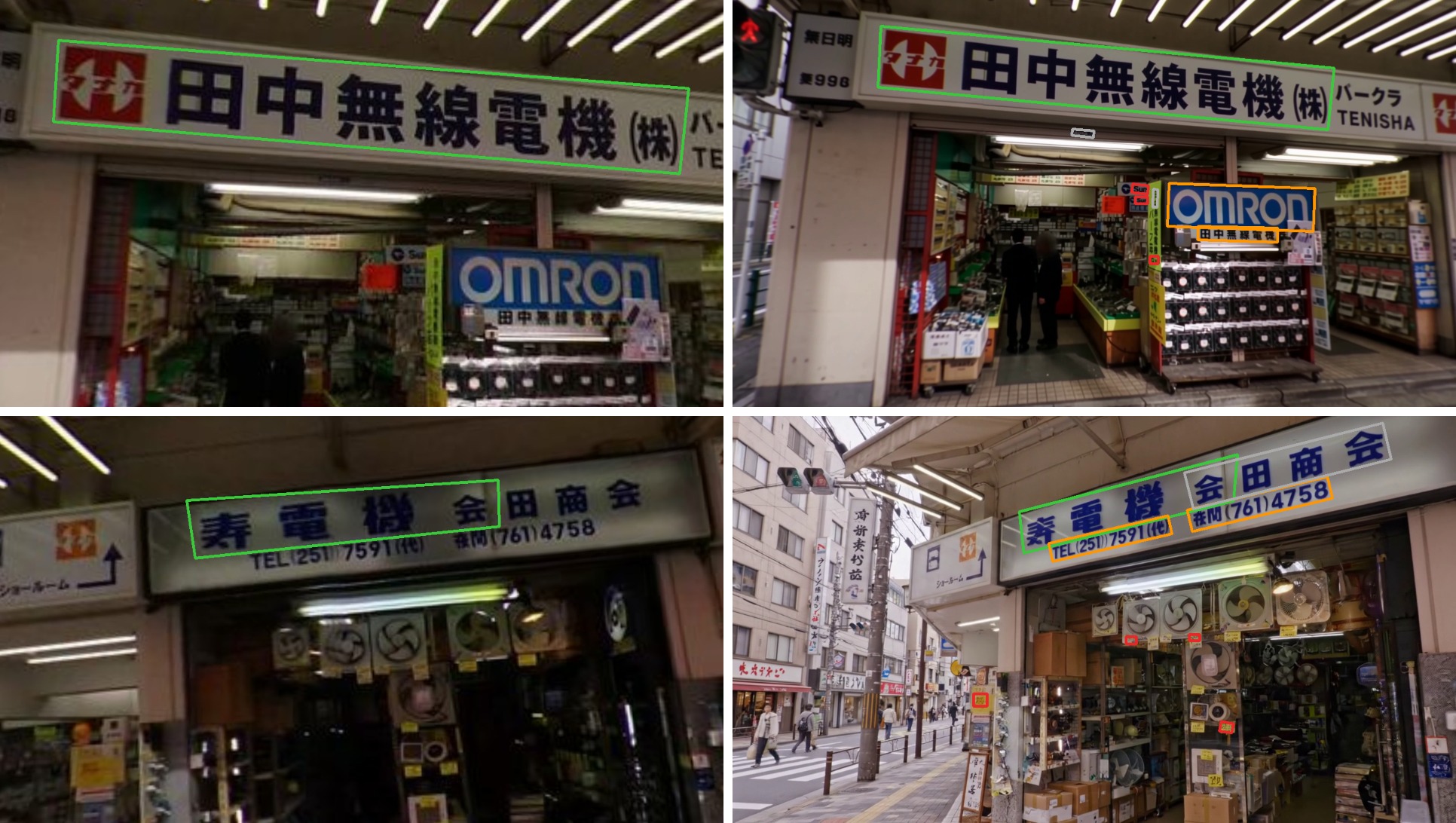}
\caption{Reference (left) and generated frame at 85\% of the clip (right) for one held-out shot per model, chosen by rule under R1: among held-out shots with $\geq$ 5 scored peripheral boxes, the one whose peripheral unsupported rate is closest to the model's (Wan 3.0, shot 42: 3 of 7, model 42.0\%; Veo 3.1 Fast, shot 43: 4 of 7, model 55.5\%). Green: anchor. Red: peripheral box unsupported by the reference. Orange: sharpened or altered. Gray: supported. The figure shows the measure's noise as it is: the two red ``Sun'' boxes (top) mark a sticker that is present in the reference but was not detected there, and the orange boxes are matching errors of the altered category (a case difference, and boxes paired with a neighbouring reference string), which is not part of the headline rate. Mapillary images 1411229289226379 and 153100413444387 (potaro67v), re-projected and cropped, CC BY-SA 4.0.}\label{fig:1}
\end{figure*}

A model that is shown a real place should give back that place. As video generation moves into applications that show real locations --- navigation previews, virtual tours, simulation --- a specific failure becomes consequential: the model receives a photograph of a \emph{specific} street and returns a video of a \emph{plausible} one. In the generations we measured, the shop sign the prompt is about comes back intact. Around it, menu boards, posters and neighbouring signs fill with crisp, confident text that the photograph does not contain (Figure 1).

This failure is invisible to current evaluations. Compositional benchmarks score agreement with the \emph{prompt} \cite{sun2025t2vcompbench,liu2024evalcrafter}. VBench++'s image-to-video track does compare against the conditioning image \cite{huang2024vbenchpp}, but through global feature similarity, which certifies that the scene still looks like the same scene, not that a given string still reads the same; on our videos its two scores show no evidence of tracking the fabrication we report (\S{}5.5). Text benchmarks that start from a reference frame score text content on the annotated target block \cite{zhang2026beyondlegibility}; other text is checked for malformed strokes, not against the reference. Outside video editing, where source text beyond the edit target is checked for preservation \cite{zhang2026beyondlegibility}, none of the generation evaluations we examined measures the setting such applications create: \textbf{the model is given the reference, and every string of it in the frame --- not only the one being checked --- must survive generation.}

The natural way to measure this is to annotate the important text in the reference --- here, the destination sign --- track it through the video, rectify it and read it. We built exactly such a metric, anchor text fidelity (ATF, \S{}4.2), and validated it against human judgment. On it, four models from three vendors sit close to a no-model ceiling and are not reliably separated. Measured over the periphery --- the text elsewhere in the frame that the reference covers, in shots that can be tracked --- with a region-level test calibrated against a no-model control (\S{}4.3), the picture inverts: each model produces substantially more unsupported text than the control (Table~\ref{tab:1}), and two blinded raters, judging crops with the control hidden among them, agree (\S{}5.2).

The anchor is not an arbitrary place to look. It is usually the largest and clearest string in the frame and the one the prompt names, and in our data it is where fidelity is highest. Removing the naming sentence did not lower anchor fidelity by the preregistered margin; text size explains part of the gap but could not be tested as an explanation for all of it (\S{}5.5). Either way, anchor-based evaluation measured the best-preserved text in the frame, not a representative sample of it.

We also report how we nearly missed this. Our first preregistered peripheral measure compared OCR strings and failed its own control, because most fabricated text is read by OCR with low confidence and a confidence gate discards it; the region measure that replaced it later failed its floor on Latin script and was revised and re-validated on held-out data before use (\S{}5.3). Predictions were registered before the data that tested them, and every model-level claim separates development data from held-out rounds.

\textbf{Contributions.} 1. \textbf{Finding.} Anchor-based evaluation misses most of the measurable text fabrication in image-conditioned video: anchor fidelity is near a no-model ceiling and does not rank four models from three vendors, while 36--54\% of trackable, reference-covered peripheral text regions are unsupported (1.6\% for the control) --- confirmed by two blinded raters, held in preregistered held-out rounds on two vendors, with no evidence of tracking by VBench-I2V in this sample, only partly attributable to localization error (up to 10--23\% of Wan's unsupported boxes), and present under a prompt that forbids new readable text (\S{}5.2, \S{}5.5). 2. \textbf{Measurement.} A region-level, recognition-free fabrication measure with a no-model floor, validated against human judgment and reported as a conservative estimate with a measured no-model floor and known false positives, with the two failures that shaped it --- of a string-level alternative and of its own first version on Latin script --- and an account of what it cannot score (\S{}4.3, \S{}5.3). 3. \textbf{Release.} Code, annotations, per-video scores, human ratings, and the fifteen preregistrations for the benchmark experiments.

\section{Related Work}

\textbf{Reference-based evaluation of video generation.} Compositional and quality benchmarks score agreement with the prompt or internal coherence \cite{sun2025t2vcompbench,liu2024evalcrafter,huang2024vbench}; a video can be smooth and consistently wrong about the place it depicts. The closest prior metric compares a video with its conditioning image: VBench++'s image-to-video subject and background consistency \cite{huang2024vbenchpp}, computed as global embedding similarity (DINO \cite{caron2021dino}, DreamSim \cite{fu2023dreamsim}). Such a proxy is not built to tell whether a string still reads the same, and on our videos it does not track peripheral fabrication (\S{}5.5). Subject-to-video suites score the identity of people and objects \cite{yuan2025opens2v}; geographic probes test what models know about places from text prompts alone \cite{liu2026gap,tan2026geofidelity}. We study the conditioned setting, in which the place is given.

\textbf{Text benchmarks score the text they asked for.} Text-to-image text benchmarks compare OCR output with the strings named in the prompt. Some read only where text was requested: AnyText-benchmark crops each line at its specified position, and its authors note that ControlNet ``tends to randomly generate pseudo-text in the background'', which their metric, focusing ``only on the specified text generation areas'', does not penalize \cite{tuo2024anytext}; VTR-Bench transcribes only the specified carriers in text-to-video \cite{huang2026vtrbench}. Others read the whole image but score recall of the target \cite{tai2025textcrafter,geng2025xomni}, so unrequested text costs nothing. MARIO-Eval and OneIG-Bench do charge for extra text, against the prompt \cite{chen2023textdiffuser,chang2025oneigbench} --- possible because there the prompt is the whole specification. With a reference image, text the prompt did not name is expected, and only a comparison with the reference can say whether it is supported. The one whole-frame check against a visual reference we found is in restoration: concurrent work on diffusion super-resolution scores text-spotting precision against a fully annotated, pixel-aligned ground truth, and also finds hallucination concentrated in small text \cite{hu2026tgsr}. In image-to-video generation the camera moves and the reference covers only part of the frame, so regions must be mapped back and scored only where the reference reaches. The closest work is the concurrent VidScribe benchmark \cite{zhang2026beyondlegibility}, whose image-to-video task likewise anchors text in a starting frame and finds that conditioning preserves text best of its tasks. Its content score uses annotated target blocks; its glyph score inspects background text for malformed strokes, not against the reference; its background score masks text; and its sources were screened to exclude extra text. Its video-editing task adds a source-text preservation check on text outside the edit target, the closest analogue to our question, but for declared source text under editing rather than unrequested text in reference-conditioned generation. Video text-editing benchmarks read text only inside the edit mask \cite{chen2026vitexbench}. In reference-conditioned generation, fabricated text that is well formed and legible passes each of these checks.

\textbf{Generation, consistency and verification.} Video text generation and editing methods \cite{liu2024textanimator,liu2025videotextpreservation,wu2025hunyuanvideo15,zeng2026steervte} render or edit text named in a prompt; none measures whether text already present in a real reference survives. World-model benchmarks score internal consistency rather than external truth \cite{duan2025worldscore,gu2025geco,wu2026pdibench,fan2026geot2vbench}; cf.~\cite{yang2025thinkinginspace,yang2025cambrians}, and inference-time verifiers check prompt alignment \cite{ma2025inferencetimescaling,lee2026videorepair,liu2025videot1}. Our result is a concrete case of the gap: a model can stay consistent with the one object it was asked about while resampling the rest of the scene.

\section{Task and Data}

\textbf{Task.} An instance is a street-level reference image, a destination anchor in it --- a quadrilateral around the destination sign with a human-verified string $g$ --- and a fixed prompt template. A model produces a video conditioned on the reference in its reference-image mode. We ask two questions of the same video. \textbf{Q1 (anchor):} does the anchor still read $g$? \textbf{Q2 (periphery):} does the text elsewhere in the frame have a counterpart in the reference? Q1 is what an anchor-based benchmark measures; our result is that a good answer to Q1 does not guarantee a good answer to Q2.

\textbf{References and anchors.} 56 storefront references from three Tokyo neighbourhoods (Nakano 43, Ameya-Yokoch\={o} 7, Akihabara 6), predominantly Japanese script with some Latin. Each is a 1280$\times$720 perspective view re-projected from a Mapillary 360\textdegree{} panorama (CC BY-SA 4.0) \cite{mapillary2026web}, framed on the destination sign; no Google Street View imagery is used. One annotator confirmed every anchor (19 quadrilaterals drawn by hand, 37 accepted from automatic proposals; strings typed or corrected from an OCR pre-fill) and recorded each anchor's \emph{reference ceiling} --- how well OCR reads it in the reference itself. Generation used the 48 references that existed when it was frozen; the other 8 appear only in the controls.

\textbf{No-model controls.} For each reference we render a 10-frame clip in which the sign's linear scale sweeps from $\surd$0.3 to $\surd$3 --- a zoom with no parallax, blur or model. Whatever a measure flags on it is the measure's own error; it is a floor, not a model baseline. The \emph{panorama} control re-projects the reference's own panorama with a narrowing field of view and is the floor for ATF. The \emph{reference-pixel} control zooms the 1280$\times$720 reference itself and is the floor for the region measure, because the panorama's extra resolution can make legible text that the reference --- and hence every model --- sees only as a blur (\S{}5.3).

\section{Measurement}

\subsection{Tracking}

On frames sampled at 2 fps, the reference is matched to the frame inside the anchor quadrilateral with SIFT \cite{lowe2004sift} (ratio test 0.75, at least 12 matches) and a RANSAC homography \cite{fischler1981ransac} (5 px). A lenient \emph{text gate} (anchor located, inlier ratio $\geq$ 0.25, $\leq$ 2\% outside the frame) decides whether the anchor is read, so that altered characters, which themselves remove inliers, cannot let an anchor escape scoring; a strict \emph{geometric gate} (inlier ratio $\geq$ 0.5, area scale in [0.3, 3]) additionally guards the appearance channel. Frames failing a gate are excluded, never scored zero; coverage is reported per model (\S{}5.2, S2).

\subsection{ATF: anchor text fidelity}

On each frame passing the text gate the anchor is rectified and scored. The OCR channel is $o(t) = 1 - \mathrm{NED}(\mathrm{OCR}(C(t)), g)$ with PP-OCRv6 \cite{zhang2026ppocrv6} on raw strings; the glyph channel $\gamma(t)$ is the cosine similarity of gradient-histogram descriptors of rectified and reference crops, on strict-gate frames only. Printed anchors use $s(t) = \tfrac12 o(t) + \tfrac12 \gamma(t)$ when both are available (the weight is fixed, not fitted), artistic anchors the glyph channel only. A video's ATF is the mean of $s(t)$ over valid frames; a video without one is unscored. We report per-model medians, overall and on printed anchors whose reference ceiling is 1.00. In a probe study with predictions registered before the held-out round (S1), ATF separated character substitution from unmodified signs almost perfectly (AUC 0.956) but tracked stroke-level erasure only weakly (held-out $\rho$ = 0.236), and its glyph channel ranked a blurred but correct sign below one with a stroke deleted. We use ATF for what it is validated on: character-level substitution at the anchor.

\subsection{Peripheral fabrication: a region-level measure}

ATF looks only inside the anchor. Outside it we score text \emph{regions}, not strings. On the frame at 85\% of the clip, every text box found by the PP-OCRv6 detector --- \textbf{with no recognition-confidence filter} --- is mapped back to the reference through the anchor homography; boxes with a short side under 8 px are dropped, and a box belongs to the anchor if its centre lies inside the mapped quadrilateral enlarged 1.15$\times$, to the periphery otherwise. Reference text is detected at four scales (1, 1.5, 2, 3$\times$, boxes pooled), because generated frames enlarge the reference and a detector run at the reference's own scale misses small text that the frame then shows (\S{}5.3). A generated box is

\begin{itemize}\item \textbf{excluded} if its back-projection is not entirely inside the reference --- there is no ground truth for it;
\item \textbf{unsupported} if no reference text box intersects its back-projection after a 26-px tolerance --- the model drew text where the reference shows none;
\item \textbf{sharpened} if the matching reference box is illegible (recognition < 0.50) and the generated one legible ($\geq$ 0.80), and \textbf{altered} if both are legible and the strings differ; both are reported, not headlined.
\end{itemize}

The tolerance is the 95th percentile of the gap between boxes that read the same high-confidence string in reference and frame, calibrated on development data. Every choice biases \emph{against} finding fabrication --- excluded edges, any-overlap support, the larger tolerance --- and raters judge about half of the regions the measure accepts to be fabricated as well (\S{}5.2), so we read \textbf{the unsupported rate as a conservative estimate}: biased toward missing fabrication, though not free of false positives (\S{}7). Rates are pooled over boxes. Preregistered decisions use 95\% bootstrap intervals from 2,000 shot resamples, paired when anchor and periphery come from the same shots. This version reports point estimates; the intervals behind each registered decision are produced by the released scripts, and the release README maps each analysis to its code and outputs.

\subsection{Human judgment of regions}

The eye is used as a second instrument, calibrated the same way. Raters see the same location cropped from reference and frame, region outlined, and choose: consistent / changed characters / text where the reference has none or is unreadable / pseudo-glyphs / cannot judge. The middle three count as \textbf{fabricated}; ``cannot judge'' leaves the denominator. Items are drawn across sets and across all automatic categories, at most three per shot and region, with control pairs mixed in blind and repeated items; raters never see the set, the region type or the automatic label. Two raters judged every pack independently.

\section{Experiments}

\subsection{Setup}

\begin{table}[t]
\centering
\footnotesize
\setlength{\tabcolsep}{3pt}
\begin{tabularx}{\linewidth}{>{\raggedright\arraybackslash}l>{\raggedright\arraybackslash}X>{\raggedright\arraybackslash}X>{\raggedright\arraybackslash}X}
\toprule
\textbf{Model} & \textbf{Access} & \textbf{Output} & \textbf{Seed} \\
\midrule
Seedance 2.0 / 2.0-mini & ByteDance Ark API & 1280$\times$720, 24 fps, 5 s & per shot, shared \\
Wan 3.0 & Alibaba Bailian API & 1280$\times$720, 30 fps, 5 s & same seeds \\
Veo 3.1 Fast & Google Gemini API & 1280$\times$720, 24 fps, 8 s & not accepted \\
\bottomrule
\end{tabularx}
\end{table}

Every model ran in reference-image mode with the same structured prompt: a scene-anchoring paragraph, a sentence naming the destination sign, a camera timeline, and a negative list that includes \emph{``Do not create new readable words, shop names, logos, captions, labels, or translated signage. If text cannot be preserved exactly from the reference, keep it blurred or unreadable.''} All 192 requests carried it. One sample was drawn per shot. Veo differs in four respects --- no seed, 8-second clips, possible server-side prompt rewriting, and audio that cannot be disabled --- so its absolute rates are not directly comparable with the others.

\textbf{Development and held-out data.} The region measure was designed and calibrated on the two Seedance runs, whose peripheral rates are therefore development evidence. Wan and Veo were run after it was frozen, under preregistrations that fixed the criteria; their rates are held-out results, excluding shots inspected in smoke tests (Wan 0--15, Veo 2--3). ATF predates all four runs.

\subsection{Main result: high anchor fidelity does not guarantee fidelity in the rest of the frame}

\begin{figure*}[t]
\centering
\includegraphics[width=0.96\textwidth]{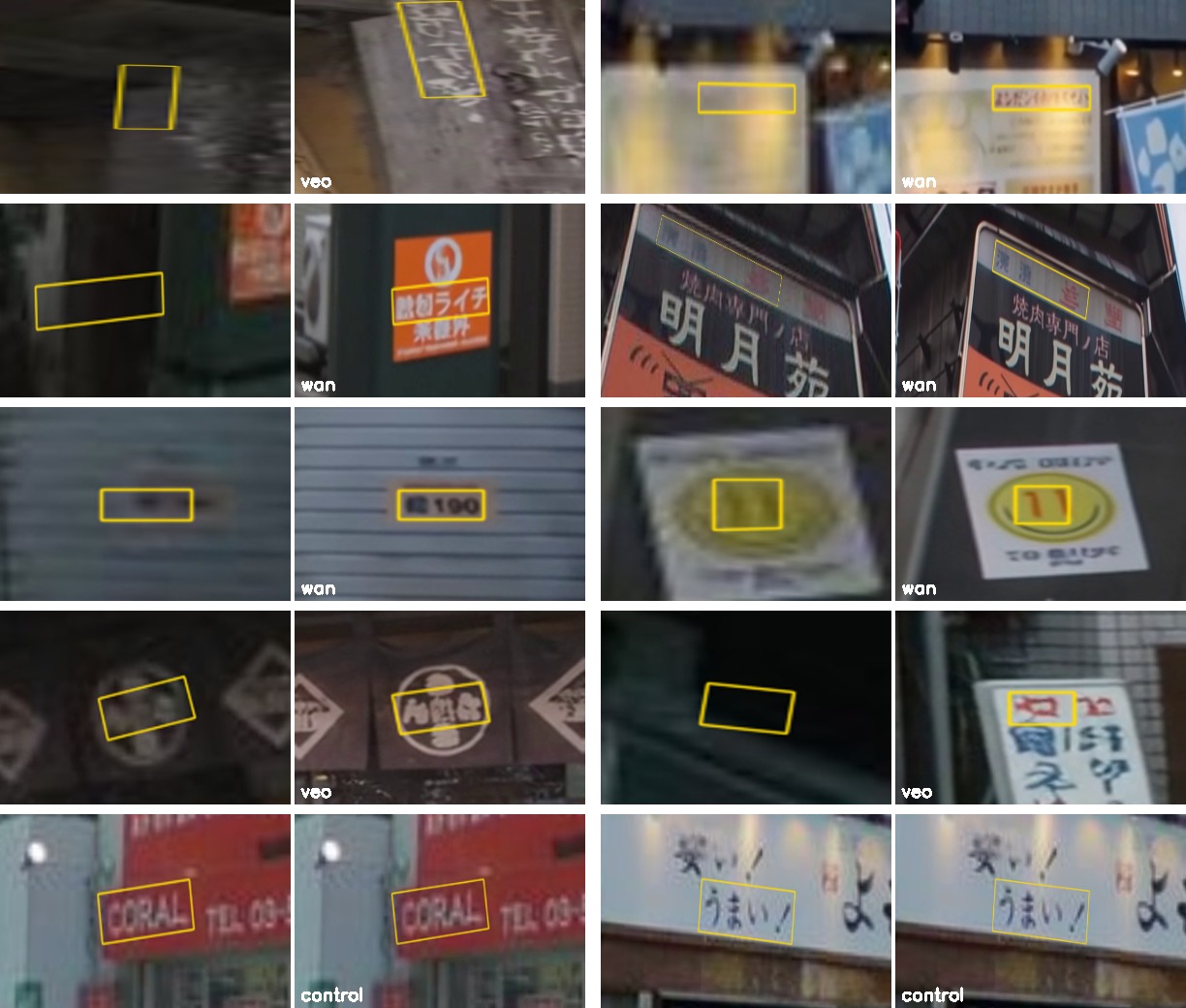}
\caption{Crop pairs exactly as shown to raters (reference left, generated right), drawn with a fixed seed: pairs both raters judged fabricated, and two control pairs.}\label{fig:2}
\end{figure*}

\begin{table*}[t]
\centering
\footnotesize
\setlength{\tabcolsep}{3pt}
\caption{Anchor fidelity (ATF, medians) and peripheral fabrication (pooled unsupported rate, boxes in parentheses). Anchor unsupported is 0.0\% in every row. Held-out rows are the registered shot sets; their gap is over the reference-pixel control.}\label{tab:1}
\begin{tabularx}{\textwidth}{>{\raggedright\arraybackslash}l>{\raggedright\arraybackslash}X>{\raggedright\arraybackslash}X>{\raggedright\arraybackslash}X>{\raggedright\arraybackslash}X>{\raggedright\arraybackslash}X}
\toprule
 & \textbf{Data} & \textbf{ATF, valid frames} & \textbf{ATF, ceiling = 1.00} & \textbf{Peripheral unsupported} & \textbf{Periphery $-$ control} \\
\midrule
Control, panorama & ATF floor & 0.996 & 0.998 & 0.0\% (61) &  \\
Control, reference pixels & region floor & --- & --- & 1.6\% (63) &  \\
Seedance 2.0 & development & 0.930 & 0.940 & 35.8\% (148) &  \\
Seedance 2.0-mini & development & 0.903 & 0.929 & 50.4\% (226) &  \\
Wan 3.0 & all shots & 0.941 & 0.965 & 42.4\% (198) &  \\
Wan 3.0 & held out (32 shots) &  &  & 42.0\% (131) & +40.4 \\
Veo 3.1 Fast & all shots & 0.928 & 0.965 & 54.1\% (135) &  \\
Veo 3.1 Fast & held out (46 shots) &  &  & 55.5\% (128) & +53.9 \\
\bottomrule
\end{tabularx}
\end{table*}

\textbf{The finding is the gap between columns, not between rows.} At the anchor every model is within 0.10 of the control, and ATF does not reliably rank them --- consistent with a registered null result for Seedance 2.0 against 2.0-mini. In the same frames, the trackable, reference-covered peripheral unsupported rate exceeds the control's 1.6\% for every model (Table~\ref{tab:1}). On the held-out vendors the periphery exceeds the control by 40.4 percentage points for Wan and 53.9 for Veo, and every registered criterion held. The gap shrinks as the support tolerance widens but stays above zero: at 0 and 77 px the held-out gaps over the control are +51.1 and +17.5 for Wan and +64.0 and +33.6 for Veo, so at three times the frozen tolerance Wan falls below the registered 20-point margin (S6). This happens under a prompt that forbids it: all four models complied at the named sign and not around it. Figure 1 shows a typical case per held-out model, selected by a fixed rule.

\textbf{Coverage.} Two losses shape these rates. Veo loses 17 of 48 shots to tracking, because in reference-image mode it re-synthesises the scene beyond what a planar homography absorbs; if a model's worst fabrications are what break tracking, its rate is underestimated. And models frame wider than the reference: text in the extra field of view has no ground truth and is excluded --- for Wan more boxes are excluded (375) than scored (281). Exclusion can only remove fabrication from the count, but it means the rates describe the part of each frame the reference covers (S2).

\begin{table*}[t]
\centering
\footnotesize
\setlength{\tabcolsep}{3pt}
\caption{Human judgment of regions: fabrication rate among judgeable peripheral regions for each rater, and agreement on fabricated-or-not.}\label{tab:2}
\begin{tabularx}{\textwidth}{>{\raggedright\arraybackslash}l>{\raggedright\arraybackslash}X>{\raggedright\arraybackslash}X>{\raggedright\arraybackslash}X}
\toprule
\textbf{Round} & \textbf{Control, r1 / r2} & \textbf{Model periphery, r1 / r2} & \textbf{$\kappa$} \\
\midrule
Seedance 2.0 (development) & 0\% / 4\% & 52\% / 56\% & 0.83 \\
Seedance 2.0-mini (development) & (same pack) & 76\% / 77\% &  \\
Wan 3.0, held out & 0\% / 0\% & 70\% / 60\% & 0.91 \\
Veo 3.1 Fast, held out & 0\% / 9\% & 81\% / 79\% & 0.96 \\
\bottomrule
\end{tabularx}
\end{table*}

\textbf{Raters agree.} In every round each rater flagged a greater share of judgeable peripheral regions as fabricated on model outputs than on the control. Table~\ref{tab:2} gives the rates and binary agreement for the main rounds. On the held-out rounds both raters found positive periphery-minus-anchor gaps under the shot-bootstrap check. The point estimates were +54 and +51 percentage points for Wan and +64 and +62 for Veo. Of the regions the automatic measure called unsupported, raters judged 90--100\% fabricated, against 25--55\% of those it called supported: the automatic rate is valid and conservative. These packs were sampled with the measure's first version (R0, \S{}5.3).

\textbf{What the fabrication looks like.} Across all rounds and both raters, two peripheral regions in total were judged a changed-character substitution, and none at any anchor: models do not rewrite neighbouring shop names into other names. Fabricated regions are text where the reference shows none or cannot be read, or pseudo-glyphs (Figure 2). Raters often disagreed on which of the two while agreeing that a region was fabricated, so we report only the union. Both forms leave the reference no legible string to compare against --- which is what defeats string-based measures.

\subsection{How the measure was built: two failures}

\textbf{A string-level measure failed its control.} Our first preregistered attempt read every text line with OCR, kept lines with recognition confidence $\geq$ 0.80, and counted a generated string as corrupted when it was near but not equal to a reference string. On the control the periphery already exceeded the anchor by 6 points --- small text is misread more often --- and the model gaps (+5, +1) did not exceed that, so the measure was withdrawn as registered. Most fabricated text is read with low confidence, so the gate removed the very failure it was meant to count. A separate two-rater calibration of OCR-based labels on other data (Appendix~S12) is consistent with this: the labeller recovered only about a quarter of the outputs humans judged unreadable. The region measure drops recognition entirely.

\textbf{The region measure failed its floor on Latin script.} To test generality we preregistered 30 storefronts in Soho, London, frozen before generation, with the control as a gate. Under the region measure as it then stood (R0: reference text detected at its own scale only, panorama control), the control flagged 8 of 58 peripheral regions (13.8\%), above the registered 5\%, and no model was run. None of the eight was fabrication: three were graphics detected as text, the rest reference text the detector had missed at the reference's scale. On these 30 shots --- development data, used to choose the fix --- detecting reference text at four scales against the reference-pixel control left 0 of 60 regions flagged.

\textbf{The revision was validated before use.} Under a preregistration written before it was run on any of them, the revised measure (R1, \S{}4.3) flagged 0 of 54 regions on the 41 London references not selected for generation and 1 of 63 on the 56 Tokyo references, both within the floor (R0: 7/54 and 3/63). Only then were all four Tokyo models re-scored on their existing videos. Every model rate fell, and every registered criterion still held; Table 1 reports R1 and the supplement the original numbers (S6). The revision errs conservatively: finding more reference text can only turn unsupported regions into supported ones, including some truly fabricated.

\subsection{Latin script: London}

\begin{table*}[t]
\centering
\footnotesize
\setlength{\tabcolsep}{3pt}
\caption{London, R1, against the London held-out floor (0/54). Peripheral unsupported rates (numbers of scored boxes in parentheses); anchor unsupported is 0.0\% throughout.}\label{tab:3}
\begin{tabularx}{\textwidth}{>{\raggedright\arraybackslash}l>{\raggedright\arraybackslash}X>{\raggedright\arraybackslash}X}
\toprule
 & \textbf{Wan 3.0} & \textbf{Veo 3.1 Fast} \\
\midrule
Round A (30 references) & 24.8\% (149) & 19.8\% (96) \\
Round B (40 references) & 32.8\% (177) & 36.8\% (117) \\
Pooled (descriptive) & 29.1\% (326) & 29.1\% (213) \\
\bottomrule
\end{tabularx}
\end{table*}

With the floor passed, both held-out vendors were run on the 30 frozen London references. The registration required a periphery at least 20 points above the floor and above the same shots' anchors, with intervals above zero. Wan met both; Veo missed both by 0.2 points, so \textbf{round A did not replicate}, and we keep that verdict. Because round A was underpowered, we preregistered a second, independent round on the 40 remaining qualified references, with the same models, settings and thresholds and no third round allowed; \textbf{round B met every criterion for both models.} We therefore describe the Latin-script result as replicated in the second of two preregistered rounds, not outright; the pooled row is descriptive. A registered human check of the London regions agreed with the measure --- raters judged 96\% and 100\% of unsupported regions fabricated against 54\% and 58\% of supported ones ($\kappa$ = 0.91), and at most 1 of the 26 unsupported regions was marked ``cannot judge'' by either rater --- the answer that covers non-text --- but one rater judged 5 of 15 control regions fabricated, failing the registered human floor, so this check is descriptive only (S7).

\subsection{Alternative explanations}

\textbf{Is the anchor kept because it is named?} We regenerated all 48 Wan shots with the same seeds and a prompt identical except for the sentence naming the sign, keeping every general fidelity instruction. The videos changed (median per-shot pixel difference 25.6/255), but anchor fidelity did not: the paired median ATF difference was 0.000 over 38 shots, and the anchor and peripheral unsupported rates moved by 0.0 and +0.1 points, all below the registered thresholds. Naming alone is not what protects the anchor in this experiment; salience was not manipulated, and the general instructions may protect it on their own.

\textbf{Is it only text size?} Anchor text is large (median short side 66.5 px on Wan's held-out shots, 120 px on Veo's) and peripheral text small (24 px). After reweighting peripheral boxes to the anchor's size distribution, the held-out gaps were 15.9 percentage points for Wan and 21.8 for Veo. Both remained positive under the shot-bootstrap check, but the registered test is \textbf{undecidable}: it needs 20 peripheral boxes at least as large as the anchor's median, and there are 17 (Wan) and 7 (Veo). For Wan, large peripheral text is almost always supported (2.2\% unsupported at $\geq$ 32 px, 1 of 46 boxes), so its gap is nearly all a small-text effect; Veo fabricates at the anchor's scale (30.2\%). Size and anchor status are confounded in this data (S4).

\textbf{Does an existing reference-based metric already see it?} We ran VBench-I2V's two dimensions with the official implementation unchanged on all 248 videos and asked whether each shot's score tracks its peripheral unsupported rate $f$ (66 shots with $\geq$ 3 boxes).

\begin{table*}[t]
\centering
\footnotesize
\setlength{\tabcolsep}{3pt}
\caption{VBench-I2V against peripheral fabrication: Spearman $\rho$ with $f$ (negative if the metric sees fabrication) and AUC ranking low- ($f \le 0.2$, n = 11) above high-fabrication shots ($f \ge 0.5$, n = 40).}\label{tab:4}
\begin{tabular*}{\textwidth}{@{\extracolsep{\fill}}lrrrrrrr@{}}
\toprule
\textbf{Dimension} & \textbf{$\rho$(score, $f$)} & \textbf{AUC} & \textbf{Control} & \textbf{Seedance 2.0} & \textbf{2.0-mini} & \textbf{Wan 3.0} & \textbf{Veo 3.1 Fast} \\
\midrule
Subject consistency & +0.122 & 0.386 & 0.9514 & 0.9486 & 0.9436 & 0.8620 & 0.8745 \\
Background consistency & +0.090 & 0.375 & 0.9614 & 0.9608 & 0.9561 & 0.8943 & 0.9016 \\
\bottomrule
\end{tabular*}
\end{table*}

Both dimensions met the registered criterion for ``does not track'', which is a criterion, not a test of equivalence; the point estimates even run the wrong way, and Seedance 2.0 scores within 0.003 of the no-model control while half its periphery (R0) is unsupported. With 11 low-fabrication shots, what is excluded is strong separation, not weak.

\textbf{Is ``unsupported'' just misalignment?} Two checks avoid the anchor homography (both under R0). \emph{Local re-registration} fits a homography around each box: on Wan's held-out shots 83\% of boxes re-register, the unsupported rate stays at 41.3\% against 1.7\% on the control, and 90\% of unsupported boxes stay unsupported; for Veo only 48\% re-register, below the registered minimum, so the check is undecidable. In a \emph{position-free human search}, raters saw the whole reference beside a generated box and said whether its text appears anywhere: for Veo they found 10\% and 14\% of unsupported boxes elsewhere, within the registered 20\%; for Wan, 29\% and 17\%. In Wan's full set of 30 sampled unsupported boxes, both raters found the text elsewhere for 4 boxes and at least one found it for 7 (S5).

\section{Discussion}

\textbf{Score the frame, not the anchor.} The region a benchmark annotates is almost by construction the text a conditioned model is most likely to preserve --- the largest, the clearest, and in a task prompt the one named. Our own primary metric inherited this blind spot: on it four models from three vendors cluster near the ceiling without a reliable ranking, while over the measurable periphery every model has a substantially higher unsupported rate than the control. The consequence does not depend on the mechanism: a fidelity score computed only where the evaluator looked overstates fidelity, and should be reported alongside a frame-level measure with a no-model floor and an account of what could not be scored.

\textbf{What the fabrication looks like decides what can see it.} The fabricated text is almost never a substituted string; it is text where the reference shows none or cannot be read. String metrics drop it at a confidence gate or file it as unknown, and the first measure we tried failed for exactly this reason. Region-level detection against a control, checked by human judgment, is the minimum we found that works.

\textbf{For applications and generation.} A system that verifies only the destination sign can pass videos with substantial unsupported text around it, even when the prompt forbids it. Verifiers used to filter or guide generation will need to look beyond the anchor; if the anchor's advantage is largely scale, constraining or verifying small text, or blurring text that cannot be verified, are natural directions we did not test.

\section{Limitations and Ethics}

\textbf{Scale and scope.} The main result rests on 48 shots per model from three Tokyo neighbourhoods, one sample per shot, four models from three vendors; it is an existence-and-magnitude result, not a ranking. The Seedance rates are development data; only Wan and Veo are held out. Beyond Japanese script the evidence is two rounds in one London neighbourhood, the first of which fell short.

\textbf{Measurement.} The region measure is conservative by design but not a guaranteed lower bound, since it has false positives: for Wan 10--23\% of unsupported boxes may be localization error, and in the position-free search 20--30\% were judged not to be text (S5). Its magnitude depends on the support tolerance. It was revised after a failure and chosen on the failed set; its independent checks are two held-out floors, and the human packs, local re-registration and VBench analysis were run under the original rule. It scores only what the reference covers and loses shots the tracker cannot follow (up to 17 of 48 for Veo). Where the reference is blurred, ``unsupported'' means unsupported by the input, not wrong. ATF is weak on stroke-level damage. Occlusion is not gated, and one person annotated the anchors. The string comparison behind the \emph{altered} category is case-sensitive, so Latin text that differs only in case is counted as altered; altered is reported but not part of any headline rate.

\textbf{Raters.} Two raters judged every pack: one is the author; the other is a classmate with no stake in the work, who took part unpaid and with informed consent. The same two raters performed the calibration in S12. Control pairs are near-identical and may be recognisable, so a low control rate shows only that raters do not invent differences; agreement is high on fabricated-or-not and lower on fabrication type, so only the former is reported.

\textbf{Ethics and licensing.} References are Mapillary images under CC BY-SA 4.0, re-projected and cropped, with per-image attribution in the release; they carry Mapillary's automatic face and licence-plate blurring, and every released crop and figure was screened with a face detector and the flags reviewed by hand (none showed an identifiable face in generated content). Ratings were collected from consenting adults on image crops only, with no personal data. Generated videos are released only where provider terms allow; otherwise we release per-video scores and the features needed to reproduce every table.

{\small
\bibliographystyle{ieeenat_fullname}
\bibliography{references}

@inproceedings{sun2025t2vcompbench,
  title     = {{T2V-CompBench: A Comprehensive Benchmark for Compositional Text-to-video Generation}},
  author    = {Sun, Kaiyue and Huang, Kaiyi and Liu, Xian and Wu, Yue and Xu, Zihan and Li, Zhenguo and Liu, Xihui},
  booktitle = {Proceedings of the IEEE/CVF Conference on Computer Vision and Pattern Recognition (CVPR)},
  year      = {2025},
  pages     = {8406--8416},
  note      = {arXiv:2407.14505}
}

@inproceedings{liu2024evalcrafter,
  title     = {{EvalCrafter: Benchmarking and Evaluating Large Video Generation Models}},
  author    = {Liu, Yaofang and Cun, Xiaodong and Liu, Xuebo and Wang, Xintao and Zhang, Yong and Chen, Haoxin and Liu, Yang and Zeng, Tieyong and Chan, Raymond and Shan, Ying},
  booktitle = {Proceedings of the IEEE/CVF Conference on Computer Vision and Pattern Recognition (CVPR)},
  year      = {2024},
  pages     = {22139--22149},
  note      = {arXiv:2310.11440}
}

@inproceedings{huang2024vbench,
  title     = {{VBench: Comprehensive Benchmark Suite for Video Generative Models}},
  author    = {Huang, Ziqi and He, Yinan and Yu, Jiashuo and Zhang, Fan and Si, Chenyang and Jiang, Yuming and Zhang, Yuanhan and Wu, Tianxing and Jin, Qingyang and Chanpaisit, Nattapol and others},
  booktitle = {Proceedings of the IEEE/CVF Conference on Computer Vision and Pattern Recognition (CVPR)},
  year      = {2024},
  pages     = {21807--21818},
  note      = {arXiv:2311.17982}
}

@article{huang2024vbenchpp,
  title         = {{VBench++: Comprehensive and Versatile Benchmark Suite for Video Generative Models}},
  author        = {Huang, Ziqi and Zhang, Fan and Xu, Xiaojie and He, Yinan and Yu, Jiashuo and Dong, Ziyue and Ma, Qianli and Chanpaisit, Nattapol and Si, Chenyang and Jiang, Yuming and others},
  journal       = {arXiv preprint arXiv:2411.13503},
  year          = {2024},
  eprint        = {2411.13503},
  archivePrefix = {arXiv}
}

@article{yuan2025opens2v,
  title         = {{OpenS2V-Nexus: A Detailed Benchmark and Million-Scale Dataset for Subject-to-Video Generation}},
  author        = {Yuan, Shenghai and He, Xianyi and Deng, Yufan and Ye, Yang and Huang, Jinfa and Lin, Bin and Luo, Jiebo and Yuan, Li},
  journal       = {arXiv preprint arXiv:2505.20292},
  year          = {2025},
  eprint        = {2505.20292},
  archivePrefix = {arXiv}
}

@article{liu2026gap,
  title         = {{Are Video Generation Models Geographically Fair? An Attraction-Centric Evaluation of Global Visual Knowledge}},
  author        = {Liu, Xiao and Zhang, Jiawei},
  journal       = {arXiv preprint arXiv:2601.18698},
  year          = {2026},
  eprint        = {2601.18698},
  archivePrefix = {arXiv}
}

@article{tan2026geofidelity,
  title         = {{GeoFidelity-Bench: Evaluating Block-Conditioned Geographic Fidelity in Text-to-Image Street-View Generation}},
  author        = {Tan, Kaizhen},
  journal       = {arXiv preprint arXiv:2606.23669},
  year          = {2026},
  eprint        = {2606.23669},
  archivePrefix = {arXiv}
}

@article{liu2024textanimator,
  title         = {{Text-Animator: Controllable Visual Text Video Generation}},
  author        = {Liu, Lin and Liu, Quande and Qian, Shengju and Zhou, Yuan and Zhou, Wengang and Li, Houqiang and Xie, Lingxi and Tian, Qi},
  journal       = {arXiv preprint arXiv:2406.17777},
  year          = {2024},
  eprint        = {2406.17777},
  archivePrefix = {arXiv}
}

@article{liu2025videotextpreservation,
  title         = {{Video Text Preservation with Synthetic Text-Rich Videos}},
  author        = {Liu, Ziyang and Valencia, Kevin and Cui, Justin},
  journal       = {arXiv preprint arXiv:2511.05573},
  year          = {2025},
  eprint        = {2511.05573},
  archivePrefix = {arXiv}
}

@article{zhang2026beyondlegibility,
  title         = {{Beyond Legibility: Benchmarking Visual Text Rendering and In-Place Editing in Unified Video Generation}},
  author        = {Zhang, Ziying and Li, Litao and Liao, Junchao and Zeng, Tianyi and Zhu, Siyu and Qin, Long and Zhang, Zhenghao},
  journal       = {arXiv preprint arXiv:2609.36598},
  year          = {2026},
  eprint        = {2609.36598},
  archivePrefix = {arXiv}
}

@article{huang2026vtrbench,
  title         = {{VTR-Bench: A Systematic Benchmark for Evaluating Visual Text Rendering in Video Generation}},
  author        = {Huang, Yu and Li, Jungang and Wang, Zhiyuan and Hei, Yonghua and Dai, Song and Yang, Jiayu and Liu, Deyuan and Zheng, Xiang and Shi, Xiaoshuang and Cheng, Hao and others},
  journal       = {arXiv preprint arXiv:2610.01499},
  year          = {2026},
  eprint        = {2610.01499},
  archivePrefix = {arXiv}
}

@article{chen2026vitexbench,
  title         = {{ViTeX-Bench: Benchmarking High-Fidelity Video Scene Text Editing}},
  author        = {Chen, Xinghao and Gao, Xiangbo and Yu, Jiongze and Wu, Yuheng and Tu, Zhengzhong},
  journal       = {arXiv preprint arXiv:2609.40356},
  year          = {2026},
  eprint        = {2609.40356},
  archivePrefix = {arXiv}
}

@inproceedings{chen2023textdiffuser,
  title     = {{TextDiffuser: Diffusion Models as Text Painters}},
  author    = {Chen, Jingye and Huang, Yupan and Lv, Tengchao and Cui, Lei and Chen, Qifeng and Wei, Furu},
  booktitle = {Advances in Neural Information Processing Systems (NeurIPS)},
  year      = {2023},
  note      = {arXiv:2305.10855}
}

@inproceedings{tuo2024anytext,
  title     = {{AnyText: Multilingual Visual Text Generation And Editing}},
  author    = {Tuo, Yuxiang and Xiang, Wangmeng and He, Jun-Yan and Geng, Yifeng and Xie, Xuansong},
  booktitle = {International Conference on Learning Representations (ICLR)},
  year      = {2024},
  note      = {arXiv:2311.03054}
}

@article{hu2026tgsr,
  title         = {{TGSR: Text-Grounded Diffusion for Faithful Image Super-Resolution}},
  author        = {Hu, Zhiming and Zhang, Ran and Ye, Angela and Goyal, Raghav and Ali, Faraz and Aumentado-Armstrong, Tristan Ty and Rangrej, Samrudhdhi B. and Derpanis, Konstantinos G. and Mohomed, Iqbal and Levinshtein, Alex},
  journal       = {arXiv preprint arXiv:2610.14281},
  year          = {2026},
  eprint        = {2610.14281},
  archivePrefix = {arXiv}
}

@article{geng2025xomni,
  title         = {{X-Omni: Reinforcement Learning Makes Discrete Autoregressive Image Generative Models Great Again}},
  author        = {Geng, Zigang and Wang, Yibing and Ma, Yeyao and Li, Chen and Rao, Yongming and Gu, Shuyang and Zhong, Zhao and Lu, Qinglin and Hu, Han and Zhang, Xiaosong and others},
  journal       = {arXiv preprint arXiv:2507.22058},
  year          = {2025},
  eprint        = {2507.22058},
  archivePrefix = {arXiv}
}

@inproceedings{chang2025oneigbench,
  title     = {{OneIG-Bench: Omni-dimensional Nuanced Evaluation for Image Generation}},
  author    = {Chang, Jingjing and Fang, Yixiao and Xing, Peng and Wu, Shuhan and Cheng, Wei and Wang, Rui and Zeng, Xianfang and Yu, Gang and Chen, Hai-Bao},
  booktitle = {Advances in Neural Information Processing Systems (NeurIPS)},
  year      = {2025},
  note      = {arXiv:2506.07977}
}

@article{tai2025textcrafter,
  title         = {{Investigating Text Insulation and Attention Mechanisms for Complex Visual Text Generation}},
  author        = {Tai, Ying and Du, Nikai and Xie, Rui and Chen, Zhennan and Wang, Qian and Jiang, Zhengkai and Zhang, Kai and Yang, Jian},
  journal       = {arXiv preprint arXiv:2503.23461},
  year          = {2025},
  eprint        = {2503.23461},
  archivePrefix = {arXiv}
}

@article{wu2025hunyuanvideo15,
  title         = {{HunyuanVideo 1.5 Technical Report}},
  author        = {Wu, Bing and Zou, Chang and Li, Changlin and Huang, Duojun and Yang, Fang and Tan, Hao and Peng, Jack and Wu, Jianbing and Xiong, Jiangfeng and Jiang, Jie and others},
  journal       = {arXiv preprint arXiv:2511.18870},
  year          = {2025},
  eprint        = {2511.18870},
  archivePrefix = {arXiv}
}

@article{zeng2026steervte,
  title         = {{SteerVTE: Seamless Video Text Editing with Style and Glyph Control}},
  author        = {Zeng, Kai and Li, Moran and Wang, Zhengwei and Yu, Yingchen and Lin, Yiheng and An, Ruichuan and Lu, Ming and She, Qi and Zhang, Wentao},
  journal       = {arXiv preprint arXiv:2606.23254},
  year          = {2026},
  eprint        = {2606.23254},
  archivePrefix = {arXiv}
}

@inproceedings{yang2025thinkinginspace,
  title     = {{Thinking in Space: How Multimodal Large Language Models See, Remember, and Recall Spaces}},
  author    = {Yang, Jihan and Yang, Shusheng and Gupta, Anjali W. and Han, Rilyn and Fei-Fei, Li and Xie, Saining},
  booktitle = {Proceedings of the IEEE/CVF Conference on Computer Vision and Pattern Recognition (CVPR)},
  year      = {2025},
  pages     = {10632--10643},
  note      = {arXiv:2412.14171}
}

@article{yang2025cambrians,
  title         = {{Cambrian-S: Towards Spatial Supersensing in Video}},
  author        = {Yang, Shusheng and Yang, Jihan and Huang, Pinzhi and Brown, Ellis and Yang, Zihao and Yu, Yue and Tong, Shengbang and Zheng, Zihan and Xu, Yifan and Wang, Muhan and others},
  journal       = {arXiv preprint arXiv:2511.04670},
  year          = {2025},
  eprint        = {2511.04670},
  archivePrefix = {arXiv}
}

@inproceedings{duan2025worldscore,
  title     = {{WorldScore: A Unified Evaluation Benchmark for World Generation}},
  author    = {Duan, Haoyi and Yu, Hong-Xing and Chen, Sirui and Fei-Fei, Li and Wu, Jiajun},
  booktitle = {Proceedings of the IEEE/CVF International Conference on Computer Vision (ICCV)},
  year      = {2025},
  note      = {arXiv:2504.00983}
}

@article{gu2025geco,
  title         = {{GeCo: Evaluating Geometric Consistency for Video Generation via Motion and Structure}},
  author        = {Gu, Leslie and Hur, Junhwa and Herrmann, Charles and Zhan, Fangneng and Zickler, Todd and Sun, Deqing and Pfister, Hanspeter},
  journal       = {arXiv preprint arXiv:2512.22274},
  year          = {2025},
  eprint        = {2512.22274},
  archivePrefix = {arXiv}
}

@article{wu2026pdibench,
  title         = {{Quantitative Video World Model Evaluation for Geometric-Consistency}},
  author        = {Wu, Jiaxin and Pi, Yihao and Zhang, Yinling and Li, Yuheng and Zou, Xueyan},
  journal       = {arXiv preprint arXiv:2605.15185},
  year          = {2026},
  eprint        = {2605.15185},
  archivePrefix = {arXiv}
}

@article{fan2026geot2vbench,
  title         = {{GeoT2V-Bench: Benchmarking 3D Consistency in Text-to-Video Models via 3D Reconstruction}},
  author        = {Fan, Chenrui and Favaro, Paolo},
  journal       = {arXiv preprint arXiv:2606.24829},
  year          = {2026},
  eprint        = {2606.24829},
  archivePrefix = {arXiv}
}

@article{ma2025inferencetimescaling,
  title         = {{Inference-Time Scaling for Diffusion Models beyond Scaling Denoising Steps}},
  author        = {Ma, Nanye and Tong, Shangyuan and Jia, Haolin and Hu, Hexiang and Su, Yu-Chuan and Zhang, Mingda and Yang, Xuan and Li, Yandong and Jaakkola, Tommi and Jia, Xuhui and others},
  journal       = {arXiv preprint arXiv:2501.09732},
  year          = {2025},
  eprint        = {2501.09732},
  archivePrefix = {arXiv}
}

@inproceedings{lee2026videorepair,
  title     = {{Self-Correcting Text-to-Video Generation with Misalignment Detection and Localized Refinement}},
  author    = {Lee, Daeun and Yoon, Jaehong and Cho, Jaemin and Bansal, Mohit},
  booktitle = {Findings of the Association for Computational Linguistics: ACL 2026},
  year      = {2026},
  pages     = {36464--36489},
  note      = {arXiv:2411.15115}
}

@inproceedings{liu2025videot1,
  title     = {{Video-T1: Test-Time Scaling for Video Generation}},
  author    = {Liu, Fangfu and Wang, Hanyang and Cai, Yimo and Zhang, Kaiyan and Zhan, Xiaohang and Duan, Yueqi},
  booktitle = {Proceedings of the IEEE/CVF International Conference on Computer Vision (ICCV)},
  year      = {2025},
  pages     = {18671--18681},
  note      = {arXiv:2503.18942}
}

@article{zhang2026ppocrv6,
  title         = {{PP-OCRv6: From 1.5M to 34.5M Parameters, Surpassing Billion-Scale VLMs on OCR Tasks}},
  author        = {Zhang, Yubo and Wang, Xueqing and Lin, Manhui and Zhang, Yue and Deng, Penglongyi and Sun, Ting and Gao, Tingquan and Zhang, Zelun and Liu, Jiaxuan and Zhou, Changda and others},
  journal       = {arXiv preprint arXiv:2606.13108},
  year          = {2026},
  eprint        = {2606.13108},
  archivePrefix = {arXiv}
}

@inproceedings{caron2021dino,
  title     = {{Emerging Properties in Self-Supervised Vision Transformers}},
  author    = {Caron, Mathilde and Touvron, Hugo and Misra, Ishan and Jégou, Hervé and Mairal, Julien and Bojanowski, Piotr and Joulin, Armand},
  booktitle = {Proceedings of the IEEE/CVF International Conference on Computer Vision (ICCV)},
  year      = {2021},
  pages     = {9650--9660},
  note      = {arXiv:2104.14294}
}

@inproceedings{fu2023dreamsim,
  title     = {{DreamSim: Learning New Dimensions of Human Visual Similarity using Synthetic Data}},
  author    = {Fu, Stephanie and Tamir, Netanel and Sundaram, Shobhita and Chai, Lucy and Zhang, Richard and Dekel, Tali and Isola, Phillip},
  booktitle = {Advances in Neural Information Processing Systems (NeurIPS)},
  year      = {2023},
  note      = {arXiv:2306.09344}
}

@article{lowe2004sift,
  title   = {Distinctive Image Features from Scale-Invariant Keypoints},
  author  = {Lowe, David G.},
  journal = {International Journal of Computer Vision},
  volume  = {60},
  number  = {2},
  pages   = {91--110},
  year    = {2004}
}

@article{fischler1981ransac,
  title   = {Random Sample Consensus: A Paradigm for Model Fitting with Applications to Image Analysis and Automated Cartography},
  author  = {Fischler, Martin A. and Bolles, Robert C.},
  journal = {Communications of the ACM},
  volume  = {24},
  number  = {6},
  pages   = {381--395},
  year    = {1981}
}

@misc{mapillary2026web,
  author       = {{Mapillary}},
  title        = {Street-level imagery},
  year         = {2026},
  howpublished = {\url{https://www.mapillary.com}},
  note         = {CC BY-SA 4.0; accessed 2026}
}
}

\clearpage
\onecolumn
\appendix
\section*{Appendix}
Sections S1--S12 contain the supplementary material.

\section*{S1 Validating ATF against human judgment}\label{sec:S1}\addcontentsline{toc}{section}{S1 Validating ATF against human judgment}

ATF was validated with synthetic probes before any model comparison: each of 32 reference signs was rendered unmodified, blurred, with a local erasure inside one character cell, with two adjacent characters swapped, and with one character replaced. A blinded rater scored each probe against its reference on a 1--5 scale. A target of Spearman $\rho$ $\geq$ 0.75 was set before validation.

\textbf{Development round.} One rater completed 66 judgments: five probe types for each of 12 randomly selected source signs, plus six repeated items; the remaining 20 signs were reserved. All 60 unique items received non-abstaining judgments; automatic text scores were available for 49 items from 10 signs, and the other 11 are reported as unmeasured, not scored zero.

\begin{table}[!htbp]
\centering
\footnotesize
\setlength{\tabcolsep}{3pt}
\begin{tabularx}{\linewidth}{>{\raggedright\arraybackslash}l>{\raggedright\arraybackslash}X>{\raggedright\arraybackslash}X>{\raggedright\arraybackslash}X>{\raggedright\arraybackslash}X}
\toprule
\textbf{Probe} & \textbf{Unique judgments} & \textbf{Median human score (1--5)} & \textbf{Items with text score} & \textbf{Median text fidelity (0--1)} \\
\midrule
Unmodified & 12 & 5 & 10 & 1.000 \\
Blur & 12 & 5 & 10 & 1.000 \\
Local erasure & 12 & 3 & 10 & 1.000 \\
Adjacent character swap & 12 & 1 & 9 & 0.571 \\
Character replacement & 12 & 1 & 10 & 0.633 \\
\bottomrule
\end{tabularx}
\end{table}

\textbf{Agreement is conditional on failure type, and a single $\rho$ is not a property of the metric.} Pooled over all 49 paired items, the text channel's Spearman correlation with the human rating is 0.744 (uncertainty estimated with 4,000 source-sign cluster bootstrap resamples). We report that number only to retire it: the pooled value is a property of the metric \emph{crossed with the probe mix we chose}, and the mix has no external justification.

\begin{table}[!htbp]
\centering
\footnotesize
\setlength{\tabcolsep}{3pt}
\begin{tabularx}{\linewidth}{>{\raggedright\arraybackslash}l>{\raggedright\arraybackslash}X>{\raggedright\arraybackslash}X>{\raggedright\arraybackslash}X}
\toprule
\textbf{Subset of probe families} & \textbf{$n$} & \textbf{Text $\rho$} & \textbf{Glyph $\rho$} \\
\midrule
All five (pooled) & 49 & 0.744 & 0.683 \\
Substitution vs. unmodified (swap, paste / none, blur) & 39 & 0.773 & 0.735 \\
Stroke erasure vs. unmodified (erase / none, blur) & 30 & \textbf{0.108} & 0.097 \\
\bottomrule
\end{tabularx}
\end{table}

The development correlations differ by failure type, so one pooled $\rho$ depends on the chosen probe mix. On construction labels (all probes whose label survived the erasure screen), the text channel's AUC was 0.956 for substitution versus unmodified probes and 0.565 for erasure versus blur. The glyph channel's AUC on erasure versus blur was 0.264.

Erasure probes are placed at a random position in a character cell, so some land on flat background and destroy nothing; we screen them automatically by whether the altered patch had internal structure. In development the rater separated the screen's two groups cleanly, which we registered as a prediction for the held-out round. It did not replicate (below), and we make no claim that the screen is validated against human perception.

\textbf{The 1--5 scale is reliable; a categorical second question was not, and was revised.} All six repeated fidelity ratings agreed exactly. A second question asking whether the frame contains clearly legible but incorrect text agreed on only four of six repeats and, by its wording, could not register character reordering (each character individually matches); it was revised to compare the string read in order, and the two versions are not pooled.

\textbf{Held-out round, with predictions registered before rating.} Four predictions were registered (\texttt{preregistrations/\allowbreak{}probes-\allowbreak{}validation-\allowbreak{}2026-\allowbreak{}10-\allowbreak{}01.\allowbreak{}md}) before any rating of the 20 reserved signs existed; the same rater judged 110 items (100 unique, 10 repeats) with no abstentions, and 87 unique items carried an automatic score. These signs were withheld from human rating, not from the automated probe analysis, so this is a held-out human-validation round, not an unseen test set; one rater completed both rounds, so the design does not control for a rater inferring the probe construction.

\begin{table}[!htbp]
\centering
\footnotesize
\setlength{\tabcolsep}{3pt}
\begin{tabularx}{\linewidth}{>{\raggedright\arraybackslash}l>{\raggedright\arraybackslash}X}
\toprule
\textbf{Registered prediction} & \textbf{Outcome} \\
\midrule
1. The erasure screen predicts human ratings & \textbf{Failed} \\
2. Stratified agreement stays divergent & \textbf{Half failed} \\
3. The false-alarm side stays clean & Held \\
4. Pooled $\rho$ still moves with the probe mix & Held \\
\bottomrule
\end{tabularx}
\end{table}

Prediction 1 failed (group medians 3 vs 3 rather than $\geq$ 1 apart), and the corresponding development claim is withdrawn. For prediction 2, substitution agreement was $\rho$ = 0.816 and met the registered 0.6 threshold. Erasure agreement was $\rho$ = 0.236, a weak positive association that narrowly passed the bootstrap check against zero, contrary to the prediction. Predictions 3 and 4 held: all 20 unmodified probes were rated 5 and the 20 blurred probes had median 5, with no item among those 40 rated 1 or 2; and the pooled correlation again moved with the mix --- 0.755 overall, 0.816 on substitution, 0.236 on erasure. The pooled 0.755 clears the 0.75 target, and reporting it as ``validated'' would have been the natural move; the stratified table is what we report instead. \textbf{ATF is therefore validated for character-level substitution and weak on stroke-level damage.} Every number in this subsection is produced by \texttt{refanchor\_\allowbreak{}bench/\allowbreak{}tools/\allowbreak{}human\_\allowbreak{}stats.\allowbreak{}py} and \texttt{probe\_\allowbreak{}stats.\allowbreak{}py}.

Probes (\texttt{make\_\allowbreak{}probes.\allowbreak{}py}) are built on the rectified anchor crop of each reference and pasted back into the reference's own quadrilateral, so that a probe travels the same pipeline as a generated frame (matching, rectification, OCR). The crop is split into equal character cells along its long axis, one per ground-truth character. Five probes are made per sign: \textbf{unmodified}; \textbf{blur}, a Gaussian blur with $\sigma$ = max(0.8, 1.2\% of the crop's short side), which degrades without changing content; \textbf{local erasure}, a bar one third of a cell wide and one sixth high, at a random position inside one random cell, filled with the median colour of that cell's side edges; \textbf{adjacent swap}, two neighbouring cells exchanged; and \textbf{replacement}, one cell overwritten by another cell of the same sign, so that style, lighting and stroke weight stay consistent and only the content is wrong. Swap and replacement require at least two characters. Because an erasure may land on flat background rather than on a stroke, erasures were screened before rating (\texttt{probe\_\allowbreak{}stats.\allowbreak{}py}): an erasure counts as hitting a character only if it changed at least 30 pixels and the erased patch had a grey-level standard deviation of at least 8 in the original image. The prediction that screened-out erasures would be rated as unchanged was preregistered for the validation round (\texttt{probes-\allowbreak{}validation-\allowbreak{}2026-\allowbreak{}10-\allowbreak{}01.\allowbreak{}md}, with item IDs listed before any rating).

\section*{S2 Tracking gates, coverage and the held-out rounds}\label{sec:S2}\addcontentsline{toc}{section}{S2 Tracking gates, coverage and the held-out rounds}

For each sampled frame (2 fps), the reference is matched to the frame inside the anchor quadrilateral with SIFT features, Lowe's ratio test (0.75) and a RANSAC homography (5 px reprojection threshold); at least 12 ratio-test matches are required. Two gates decide what can be scored, and their separation matters. The \textbf{strict geometric gate} requires inlier ratio $\geq$ 0.5, anchor area scale in [0.3, 3] (normalized by raster size, so that resolution differences do not masquerade as scale changes), and at most 2\% of the anchor outside the frame. The \textbf{text gate} requires only that the anchor is located (inlier ratio $\geq$ 0.25, $\leq$ 2\% outside the frame), with no area bounds: altering characters itself removes SIFT inliers, and a single strict gate would let the most altered anchors escape scoring. Frames failing a gate are excluded rather than scored zero, and per-model coverage is reported (S2). Occlusion gating by person/vehicle masks is not implemented; occluded frames are scored as they are.

\begin{table}[!htbp]
\centering
\footnotesize
\setlength{\tabcolsep}{3pt}
\caption*{\textbf{Table C.} Coverage per model.}
\begin{tabularx}{\linewidth}{>{\raggedright\arraybackslash}l>{\raggedright\arraybackslash}X>{\raggedright\arraybackslash}X>{\raggedright\arraybackslash}X>{\raggedright\arraybackslash}X>{\raggedright\arraybackslash}X>{\raggedright\arraybackslash}X>{\raggedright\arraybackslash}X}
\toprule
\textbf{Model} & \textbf{Shots} & \textbf{ATF scored} & \textbf{Valid-frame rate} & \textbf{Region measure: tracked / lost} & \textbf{Shots with peripheral text} & \textbf{Boxes scored} & \textbf{Boxes outside reference (excluded)} \\
\midrule
No-model control, panorama & 56 & 52 & 0.72 & 52 / 4 & 29 & 135 & 0 \\
No-model control, reference pixels & 56 & --- & --- & 52 / 4 & 32 & 136 & 0 \\
Seedance 2.0 & 48 & 43 & 0.80 & 44 / 4 & 27 & 228 & 114 \\
Seedance 2.0-mini & 48 & 44 & 0.78 & 41 / 7 & 31 & 314 & 67 \\
Wan 3.0 & 48 & 38 & 0.70 & 41 / 7 & 30 & 281 & 375 \\
Veo 3.1 Fast & 48 & 32 & 0.55 & 31 / 17 & 22 & 185 & 174 \\
\bottomrule
\end{tabularx}
\end{table}

Two losses shape every rate in \S{}5.2, and both are larger for some models than for others.

\emph{Shots the tracker loses.} A shot is lost when the anchor cannot be located by homography. Veo loses 17 of 48: in reference-image mode it re-synthesises the scene rather than starting from the reference, so the geometry differs from the reference by more than a planar homography can absorb. Its rows therefore rest on fewer shots (31 for the region measure, 32 for ATF). Losses could bias rates in either direction, and we cannot tell which: if a model's worst fabrications are what break tracking, its rate is underestimated.

\emph{Text outside the reference.} Models frame wider than the reference, and text they draw in the extra field of view has no ground truth, so it is excluded. For Wan more boxes are excluded (375) than scored (281). This exclusion can only remove fabrication from the count --- an excluded box can never be scored unsupported --- so it biases the peripheral rates downward, but it also means the rates describe the part of each frame the reference covers, not the whole frame.

The control's valid-frame rate (0.72) is lower than the models' because its zoom deliberately sweeps the full measurable scale range, whose ends fall outside the strict gate; this is why coverage-penalized ATF, which multiplies fidelity by valid-frame rate, is not compared across control and models.

\textbf{Wan 3.0} was registered in two stages. A first stage of 12 new shots (shots 4--15, cut for cost before any measurement) passed every criterion but rested on three shots carrying 55 of its 60 scorable peripheral boxes. The remaining 32 shots (16--47) were then generated and registered, before generation, as the primary analysis; neither the measure, the raters nor the authors had seen any of them. On those 32, the effect is spread across the set: 21 of 28 scorable shots contain peripheral text, and the three largest contribute 39\% of boxes.

\textbf{Veo 3.1 Fast} was registered the same way, with the two smoke-test shots excluded. Of the remaining 46 shots, 29 were scorable with 128 peripheral boxes over 20 shots (the three largest carry 48\%). Veo shows the most extreme form of the pattern --- the best-preserved anchor and the most fabricated periphery of the four models --- but differs from the common protocol in the four respects listed in \S{}5.1.

All registered criteria held for both vendors: periphery above the control by $\geq$ 20 points and above the anchor by $\geq$ 20 points with intervals above zero; automatic-versus-human validity (unsupported regions judged fabricated $\geq$ 60\% of the time and $\geq$ 30 points more often than supported ones, for each rater); and inter-rater $\kappa$ $\geq$ 0.6 on fabricated-or-not.

\texttt{tables/\allowbreak{}appendix\_\allowbreak{}c\_\allowbreak{}coverage.\allowbreak{}md} lists, for every shot of every set scored with R1, whether the anchor was tracked at the 85\% frame and, if so, the number of peripheral boxes counted, how many of them are unsupported, and how many generated boxes fell outside the reference and were excluded. Untracked shots are listed by number and never scored zero. The totals row of each block reproduces the corresponding rates in Table 1 and S6--S7; ATF's frame-level coverage is in Table C (S2).

\section*{S3 The string-level measure and the misidentified held-out set}\label{sec:S3}\addcontentsline{toc}{section}{S3 The string-level measure and the misidentified held-out set}

\textbf{A string-level measure failed its control.} Our first preregistered attempt (\texttt{preregistrations/\allowbreak{}peripheral-\allowbreak{}fabrication-\allowbreak{}2026-\allowbreak{}10-\allowbreak{}05.\allowbreak{}md}) read every text line with OCR, kept lines with recognition confidence $\geq$ 0.80, and counted a generated string as corrupted when it was near, but not equal to, a reference string. On the control, the periphery already exceeded the anchor by 6 points --- small text is simply misread more often --- and the model gaps (+5 for Seedance 2.0, +1 for mini) did not exceed that; per the registered rule, the measure was withdrawn. The reason became clear only later: the dominant fabrication is text the reference cannot be compared against (\S{}5.2), which OCR reads with low confidence, so the confidence gate removed the very failure it was meant to count. The region measure of \S{}4.3 drops recognition entirely.

\textbf{A misidentified held-out set.} The preregistration of the region measure (\texttt{peripheral-\allowbreak{}regions-\allowbreak{}2026-\allowbreak{}10-\allowbreak{}06.\allowbreak{}md}) named a ``held-out'' 854$\times$480 Seedance 2.0 set. After the run we found it to be the development videos downsampled, not a new sample (frame-wise mean difference $\approx$ 2.7/255, encoding noise only); a correction is appended to the registration. We report it only as a resolution check on development data --- the effect persists at lower resolution (35.8\% peripheral unsupported against 2.6\% at the anchor) --- and the Seedance rates in Table 1 remain development evidence. The control set registered alongside it was genuinely held out and passed its 10\% veto (1.6\%).

\section*{S4 Naming and text size}\label{sec:S4}\addcontentsline{toc}{section}{S4 Naming and text size}

The anchor differs from the periphery in at least two ways: it is the most salient text in the frame, and the prompt names it. This experiment manipulates only the second. We regenerated all 48 Wan shots with the same per-shot seeds and a prompt identical except for the one sentence that names the sign; the general fidelity instructions, including the negative list, were kept (\texttt{preregistrations/\allowbreak{}unname-\allowbreak{}ablation-\allowbreak{}2026-\allowbreak{}10-\allowbreak{}08.\allowbreak{}md}; decision rule in \texttt{refanchor\_\allowbreak{}bench/\allowbreak{}tools/\allowbreak{}ablation\_\allowbreak{}compare.\allowbreak{}py}, committed before generation). The median per-shot pixel difference was 25.6/255. Across the 38 shots scored in both conditions, the paired median ATF difference (named $-$ unnamed) was 0.000. Under the registered R0 rule, the differences in unsupported rates were $-$1.2 percentage points at the anchor and +0.7 at the periphery. The ATF and anchor changes met the registered criterion for no detected naming effect, which used bootstrap bounds and margins of 0.05 ATF and 10 percentage points. Under R1, the anchor difference was 0.0 points and the peripheral difference was +0.1 points. Many anchor strings were read identically with and without the sentence.

\textbf{What this does and does not show.} In this Wan experiment, removing the naming sentence produced no fidelity drop reaching the preset thresholds. It does not show that salience protects the anchor: salience was not manipulated, and the general fidelity instructions --- scene anchoring and the negative list --- were kept, so they may protect the anchor on their own. Whether scale (text large enough to survive generation is kept, smaller text is redrawn) explains the anchor's advantage is examined next. The result is from one model with one sample per shot.

\textbf{Scale, tested as far as the data allow.} Anchor text is large (median short side 66.5 px on Wan's held-out shots, 120 px on Veo's) and peripheral text small (24 px on both), so a size effect alone could produce the gap. We reweighted each set's peripheral boxes to the anchor's size distribution --- strata at the anchor's quartiles, thin strata merged --- and compared the reweighted unsupported rate with the anchor's (\texttt{preregistrations/\allowbreak{}size-\allowbreak{}strata-\allowbreak{}2026-\allowbreak{}10-\allowbreak{}10.\allowbreak{}md}; \texttt{refanchor\_\allowbreak{}bench/\allowbreak{}tools/\allowbreak{}size\_\allowbreak{}strata.\allowbreak{}py}, both committed before any size-stratified number was computed). The registered test is \textbf{undecidable} for both held-out sets: it requires at least 20 peripheral boxes at or above the anchor's median size, and there are 17 (Wan) and 7 (Veo). Descriptively, size accounts for much of the gap but not visibly all of it.

\begin{table}[!htbp]
\centering
\footnotesize
\setlength{\tabcolsep}{3pt}
\caption*{\textbf{Table G.} Peripheral unsupported rate (R1) reweighted to the anchor's text-size distribution (gap in percentage points). The registered test is undecidable for both held-out sets, as it was under R0.}
\begin{tabularx}{\linewidth}{>{\raggedright\arraybackslash}l>{\raggedright\arraybackslash}X>{\raggedright\arraybackslash}X>{\raggedright\arraybackslash}X>{\raggedright\arraybackslash}X>{\raggedright\arraybackslash}X>{\raggedright\arraybackslash}X}
\toprule
\textbf{Set} & \textbf{Anchor} & \textbf{Periphery, raw} & \textbf{Periphery, size-reweighted} & \textbf{Reweighted gap} & \textbf{Share of raw gap remaining} & \textbf{Peripheral boxes $\geq$ anchor median size} \\
\midrule
Wan 3.0 (held-out) & 0.0\% & 42.0\% & 15.9\% & 15.9 & 0.38 & 17 \\
Veo 3.1 Fast (held-out) & 0.0\% & 55.5\% & 21.8\% & 21.8 & 0.39 & 7 \\
Seedance 2.0 (dev.) & 0.0\% & 35.8\% & 16.5\% & 16.5 & 0.46 & 11 \\
Seedance 2.0-mini (dev.) & 0.0\% & 50.4\% & 19.8\% & 19.8 & 0.39 & 18 \\
Control (reference pixels) & 0.0\% & 1.6\% & 1.8\% & 1.8 & --- & 11 \\
\bottomrule
\end{tabularx}
\end{table}

On both held-out sets, peripheral text below 32 px was unsupported 53--82\% of the time. At 32 px and above, the peripheral rate was 2.2\% for Wan (1 of 46 boxes) and 30.2\% for Veo, against 0\% for the anchors in the same size group. Reweighting by size reduced the gap in every model set, leaving 38--46\% of the raw gap (Table G). The estimates depend on the few large peripheral boxes available, and both held-out tests fell short of the registered sample requirement. Counting sharpened and altered text alongside unsupported text, Veo's reweighted gap fell to 8.4 percentage points, which did not meet the shot-bootstrap criterion for a positive gap. Most of Veo's anchor errors were alterations. In these data no anchor box is smaller than 16 px. Separating size from anchor status requires references with large peripheral text. We measured text size on the generated frame.

\section*{S5 Alternative explanations in full}\label{sec:S5}\addcontentsline{toc}{section}{S5 Alternative explanations in full}

Two objections could each make the main result smaller than it looks. Both were tested under one preregistration written before any of the data below existed (\texttt{preregistrations/\allowbreak{}baseline-\allowbreak{}and-\allowbreak{}geometry-\allowbreak{}2026-\allowbreak{}10-\allowbreak{}09.\allowbreak{}md}).

\textbf{Does an existing reference-based metric already see it?} VBench++'s I2V track is the closest prior metric that compares a video \emph{against its conditioning image} (\S{}2). We ran its two I2V dimensions --- subject consistency (DINO ViT-B/16) and background consistency (DreamSim) --- on all 248 existing videos with the official implementation unchanged (frame sampling, preprocessing and the 0.4$\cdot$max + 0.3$\cdot$mean + 0.3$\cdot$min weighting), and asked whether each video's score tracks that shot's peripheral unsupported rate $f$, over the 66 shots with at least three scored peripheral boxes, pooled across the four models ($f$ under R0, as registered).

\begin{table}[!htbp]
\centering
\footnotesize
\setlength{\tabcolsep}{3pt}
\caption*{\textbf{Table 4.} VBench-I2V against peripheral fabrication. Spearman $\rho$ between score and $f$ (expected negative if the metric sees fabrication) and AUC for ranking low-fabrication shots ($f \le 0.2$, n = 11) above high-fabrication shots ($f \ge 0.5$, n = 40).}
\begin{tabular*}{\linewidth}{@{\extracolsep{\fill}}lrrrrrrr@{}}
\toprule
\textbf{Dimension} & \textbf{$\rho$(score, $f$)} & \textbf{AUC} & \textbf{Control} & \textbf{Seedance 2.0} & \textbf{2.0-mini} & \textbf{Wan 3.0} & \textbf{Veo 3.1 Fast} \\
\midrule
I2V subject & +0.122 & 0.386 & 0.9514 & 0.9486 & 0.9436 & 0.8620 & 0.8745 \\
I2V background & +0.090 & 0.375 & 0.9614 & 0.9608 & 0.9561 & 0.8943 & 0.9016 \\
\bottomrule
\end{tabular*}
\end{table}

Both dimensions met the registered criteria on the correlation and ranking tests. The correlation estimates were positive, and neither AUC reliably separated low- from high-fabrication shots. Seedance 2.0 illustrates the result: its two VBench-I2V means are within 0.003 of the no-model control, while its periphery is 50.0\% unsupported under R0 (35.8\% under R1) against the control's 1.6\%. Only 11 shots were low-fabrication, so weak separation remains possible. The within-model correlations varied in sign across small samples. On these videos, the VBench-I2V scores are insufficient as a substitute for a text-level check.

\textbf{Is ``unsupported'' just misalignment?} The region measure maps every peripheral box to the reference through the \emph{anchor's} homography. Where the facade is not planar or the camera has parallax, that mapping can be off, and a box could be called unsupported because it lands next to its true counterpart. We checked this on the held-out sets in two ways that do not use the anchor homography (both registered and run under R0).

\begin{itemize}\item \emph{Local re-registration (automatic).} Each box is re-registered with its own homography, fitted by SIFT between a window around the box (three times its size, at least 160 px) and the full reference ($\geq$ 12 ratio-test matches, $\geq$ 8 inliers), and the support test is repeated. On Wan's held-out shots 109 of 131 boxes (83\%) re-register; under local homographies the unsupported rate is 41.3\% against 1.7\% on the control, and 90\% of the anchor-unsupported boxes that re-register stay unsupported. This meets the registered criterion (persistence $\geq$ 0.80 and $\geq$ 20 points above the control). On Veo only 62 of 128 boxes (48\%) re-register, below the registered 50\% minimum, so this check is \textbf{undecidable} for Veo (persistence there was 0.79). Boxes that re-register are the less fabricated ones --- under the anchor homography 46\% (Wan) and 47\% (Veo) of them are unsupported, against 77\% and 91\% of those that do not re-register --- so these figures describe a favourable subset.
\item \emph{Position-free human search.} Two raters saw the \textbf{whole} reference, with no location marked, beside one generated text box, and answered whether that text appears anywhere in the reference (yes / no / reference too blurred to tell / not text). The answer depends on no mapping of ours. Each rater judged 101 items: 30 automatically unsupported and 10 supported peripheral boxes from each held-out set, 15 control boxes and 6 repeats, blinded to set and label.
\end{itemize}

\begin{table}[!htbp]
\centering
\footnotesize
\setlength{\tabcolsep}{3pt}
\caption*{\textbf{Table F.} Position-free search: share of boxes judged ``appears elsewhere in the reference'' (``not text'' answers excluded from the denominator).}
\begin{tabularx}{\linewidth}{>{\raggedright\arraybackslash}l>{\raggedright\arraybackslash}X>{\raggedright\arraybackslash}X}
\toprule
\textbf{Set} & \textbf{Rater 1} & \textbf{Rater 2} \\
\midrule
Control & 14/15 (93\%) & 14/15 (93\%) \\
Wan, automatically supported & 4/10 (40\%) & 3/10 (30\%) \\
\textbf{Wan, automatically unsupported} & \textbf{7/24 (29\%)} & \textbf{4/23 (17\%)} \\
Veo, automatically supported & 4/10 (40\%) & 4/7 (57\%) \\
\textbf{Veo, automatically unsupported} & \textbf{2/21 (10\%)} & \textbf{3/21 (14\%)} \\
\bottomrule
\end{tabularx}
\end{table}

The registered criterion was that at most 20\% of automatically unsupported boxes are found elsewhere, provided raters can find text that is present ($\geq$ 60\% on supported and control boxes together, met by both). For Veo it held for both raters. For Wan the raters split --- 29\% for one, 17\% for the other --- and the registration fixed no rule for combining them, so we report the split. Both raters found 4 of Wan's 30 unsupported boxes elsewhere in the reference; at least one found 7.

\textbf{What the two checks establish.} In the position-free search, raters found relatively few automatically unsupported regions elsewhere in the reference. For Wan, 10\% of re-registered unsupported boxes became supported under local homographies. In the search, both raters found 4 of 30 Wan boxes elsewhere and at least one found 7. Veo met the human-search criterion for both raters, while the local re-registration check fell below its coverage requirement. Wan's peripheral rates should therefore be read with a discount for localization error of up to 10--23\%. Many answers were ``reference too blurred to tell'' (30--47\% of unsupported boxes, and half or more of Wan's supported ones), and 20--30\% of unsupported boxes were judged not to be text at all --- false detections that the automatic measure counts. The ability to find text that was present depended mainly on the control: on model outputs, raters found 4 of 10 and 3 of 10 supported boxes for Wan, and 4 of 10 and 4 of 7 for Veo (Table F), so this test has limited power to \emph{confirm} fabrication; it bounds how much of the unsupported rate can be misalignment. Each rater agreed with their own answer on 4 of 6 repeats.

\section*{S6 The original region measure (R0) and the revision}\label{sec:S6}\addcontentsline{toc}{section}{S6 The original region measure (R0) and the revision}

R0 is \S{}4.3 with reference text detected at the reference's own scale only, and the panorama control as floor. It produced every peripheral number in this paper until S6--S7, and the human packs, local re-registration and VBench analysis were registered and run on it; the tolerance sensitivity was run on both (R1 in \S{}4.3). Its tables are kept verbatim in \texttt{tables/\allowbreak{}main\_\allowbreak{}tables.\allowbreak{}md} (R1: \texttt{tables/\allowbreak{}main\_\allowbreak{}tables\_\allowbreak{}r1.\allowbreak{}md}); the rates that changed are:

\begin{table}[!htbp]
\centering
\footnotesize
\setlength{\tabcolsep}{3pt}
\begin{tabularx}{\linewidth}{>{\raggedright\arraybackslash}l>{\raggedright\arraybackslash}X>{\raggedright\arraybackslash}X>{\raggedright\arraybackslash}X>{\raggedright\arraybackslash}X}
\toprule
\textbf{Set} & \textbf{R0 periphery} & \textbf{R1 periphery} & \textbf{R0 anchor} & \textbf{R1 anchor} \\
\midrule
Control, panorama & 1.6\% (61) & 0.0\% (61) & 0.0\% & 0.0\% \\
Control, reference pixels (R1 floor) & --- & 1.6\% (63) & --- & 0.0\% \\
Seedance 2.0 (all shots) & 50.0\% (148) & 35.8\% (148) & 2.5\% & 0.0\% \\
Seedance 2.0-mini (all shots) & 63.7\% (226) & 50.4\% (226) & 3.4\% & 0.0\% \\
Wan 3.0 (all shots) & 53.0\% (198) & 42.4\% (198) & 1.2\% & 0.0\% \\
Veo 3.1 Fast (all shots) & 67.4\% (135) & 54.1\% (135) & 0.0\% & 0.0\% \\
Wan 3.0 (held-out, 16--47) & 51.1\% (131) & 42.0\% (131) & 1.8\% & 0.0\% \\
Veo 3.1 Fast (held-out) & 69.5\% (128) & 55.5\% (128) & 0.0\% & 0.0\% \\
\bottomrule
\end{tabularx}
\end{table}

Box counts are identical because R1 changes only how reference text is found; the generated frames are detected exactly as before.

\section*{S7 Revising the measure and the London replication in full}\label{sec:S7}\addcontentsline{toc}{section}{S7 Revising the measure and the London replication in full}

\textbf{What failed.} To test generality beyond Japanese script we preregistered a Latin-script set (\texttt{preregistrations/\allowbreak{}london-\allowbreak{}crossscript-\allowbreak{}2026-\allowbreak{}10-\allowbreak{}11.\allowbreak{}md}): 30 storefronts in Soho, London, selected and frozen before any generation, with the no-model floor as a gate. Under the measure as it then stood (R0), the panorama control flagged 8 of 58 peripheral regions (13.8\%) as unsupported, above the registered 5\%, and no model was run. Inspecting the eight afterwards, none was fabrication: three were graphics detected as text (a flag, a logo, an exit icon), and the rest were text present in the reference that the detector had missed at the reference's scale but found in the enlarged control frame --- two of them legible only because the panorama has more resolution than the 1280$\times$720 reference.

\textbf{Diagnosis on the failed set (development data).} Rendering the control from the reference's own pixels still left 6/60 regions flagged, so the main cause is scale, not re-projection. Detecting reference text at 1, 1.5, 2 and 3$\times$ and pooling the boxes, with the reference-pixel control, left 0/60. These 30 shots were used to choose R1 and are not evidence for it.

\textbf{Held-out validation, preregistered before R1 was run on any of it} (\texttt{preregistrations/\allowbreak{}r1-\allowbreak{}multiscale-\allowbreak{}2026-\allowbreak{}10-\allowbreak{}14.\allowbreak{}md}). On the 41 qualified London references that were not selected for generation, and on the 56 Tokyo references, the reference-pixel control under R1 flagged 0/54 (London) and 1/63 (Tokyo) peripheral regions, both within the 5\% floor (R0 on the same two controls: 7/54 and 3/63). R1 then became the paper's measure under a reading fixed in advance: all four Tokyo models were re-scored on their existing videos, with no regeneration; both held-out vendors were re-tested against the registered criteria; and the original numbers are kept for comparison (S6). Every model rate fell and every criterion still held (Table 1). R1's error runs in the conservative direction: finding more reference text can only turn unsupported regions into supported ones, including some that are truly fabricated.

\textbf{The London replication.} With the floor passed, both held-out vendors were run on the 30 frozen London references under the generation settings of the Tokyo rounds, and scored with R1 against the London held-out floor (0/54). The registration required, for each model, a periphery at least 20 points above the floor and above the same shots' anchors, with shot-bootstrap intervals above zero.

\begin{center}
\footnotesize
\setlength{\tabcolsep}{3pt}
\begin{tabularx}{\linewidth}{>{\raggedright\arraybackslash}X>{\centering\arraybackslash}p{1.1cm}>{\raggedright\arraybackslash}p{2.25cm}>{\raggedright\arraybackslash}X}
\toprule
\textbf{Set / model (refs)} & \textbf{Tracked shots} & \textbf{Peripheral unsupported (boxes)} & \textbf{Registered L1 / L2} \\
\midrule
Round A, Wan (30) & 29 & 24.8\% (37/149) & Pass / pass \\
Round A, Veo (30) & 21 & 19.8\% (19/96) & Fail / fail \\
Round B, Wan (40) & 37 & 32.8\% (58/177) & Pass / pass \\
Round B, Veo (40) & 30 & 36.8\% (43/117) & Pass / pass \\
Pooled, Wan (70) & 66 & 29.1\% (95/326) & Descriptive only \\
Pooled, Veo (70) & 51 & 29.1\% (62/213) & Descriptive only \\
\bottomrule
\end{tabularx}
\end{center}

Anchor and floor unsupported were both 0\%, so the peripheral rate is also the point-estimate gap for L1 and L2. Each comparison required a gap of at least 20 points and a shot-bootstrap interval above zero; ``pass / pass'' records those two checks. Pooled results are descriptive because round A informed the planning of round B.

In round A, Wan met both criteria; Veo, by 0.2 points, met neither, so under the registered reading \textbf{round A did not replicate on Latin script}, and we keep that verdict rather than reinterpret the threshold. Round A covered one neighbourhood, nine Veo shots went untracked, and the estimates were imprecise. We preregistered a second round before generating it (\texttt{preregistrations/\allowbreak{}london-\allowbreak{}round-\allowbreak{}b-\allowbreak{}2026-\allowbreak{}10-\allowbreak{}15.\allowbreak{}md}): the 40 remaining qualified London references --- the ones that set the floor, after one name correction and the exclusion of one sign that was not a shop name --- with the same models, settings and thresholds, and a reading fixed in advance under which no third round would be run to change the outcome. \textbf{Round B met every criterion for both models.} We therefore describe the Latin-script result as replicated in the second of two preregistered rounds, not as replicated outright. Half of round A's references were chosen for large non-anchor text and round B's were not, so the difference between rounds is descriptive as well. The size test was undecidable in both rounds (7 and 6 peripheral boxes as large as the anchor in A, 7 and 4 in B). Unlike Tokyo, where Wan's peripheral boxes of 32 px or more were almost never unsupported (2.2\%), in London round B 28\% of Wan's boxes in that size range were unsupported, and 31\% of Veo's (descriptive).

\textbf{Human check of the London regions} (\texttt{preregistrations/\allowbreak{}london-\allowbreak{}human-\allowbreak{}2026-\allowbreak{}10-\allowbreak{}16.\allowbreak{}md}). Both rounds rested on the automatic measure alone, and Latin-script street scenes carry a false-positive source Tokyo did not --- the failed London floor was largely graphics detected as text. We therefore ran the region rating of \S{}4.4 on London, registered before any London crop was rendered: 30 peripheral and 8 anchor regions from each of the four model sets, 15 and 6 from the reference-pixel control mixed in blind, and 12 repeats, rated independently by both raters. Of the peripheral regions the measure calls unsupported, the raters judged 96\% and 100\% fabricated (24/25 and 26/26 non-abstaining), against 54\% and 58\% of those it calls supported (agreement on fabricated vs. not: $\kappa$ = 0.91). Only 1 of the 26 unsupported regions was marked ``cannot judge'' --- the answer that includes ``not text'' --- so misdetected graphics are not a large part of the London rate. The registered human floor, however, failed for one rater: on the control, one rater judged 1 of 10 non-abstaining regions fabricated and the other 5 of 15, all as pseudo-glyphs, mostly on regions the first rater declined to judge. The control here enlarges the reference's own pixels, and blurred enlarged text was read as garbled by one rater and as unjudgeable by the other. Under the registered reading the London human result is therefore descriptive, not a validation: it agrees with the automatic measure, but its own floor was not clean.

\section*{S8 Annotation procedure and per-reference attribution}\label{sec:S8}\addcontentsline{toc}{section}{S8 Annotation procedure and per-reference attribution}

\textbf{Candidates.} Panoramas were harvested from Mapillary for each neighbourhood (\texttt{harvest\_\allowbreak{}panos.\allowbreak{}py}). Text was detected on perspective crops of each panorama (\texttt{discover\_\allowbreak{}signs.\allowbreak{}py}; for London with a Latin-script rule, at least three letters), and every detection was turned into a candidate view: a 1280$\times$720 perspective re-projection framed on the detected sign, with camera position, field of view, yaw and pitch recorded.

\textbf{Human decisions.} One annotator (an author) judged every candidate in a browser tool (\texttt{pick\_\allowbreak{}destinations.\allowbreak{}py}) showing the candidate view, the OCR pre-fill and the proposed quadrilateral. For each candidate the annotator either rejected it with a reason --- not a shop name (opening hours, street names, slogans, directional panels), too narrow to read, or illegible in the reference --- or accepted it, then either accepted the proposed quadrilateral or drew one by hand around the destination sign, and typed or corrected the ground-truth string as it reads in the reference, not as the shop is known to be called. For London, an automatic audit (\texttt{pick\_\allowbreak{}destinations.\allowbreak{}audit()}) flagged accepted picks whose camera fell outside the arrival band, duplicates of the same facade, and strings that disagree with the OCR pre-fill; each flag was resolved by the same annotator. The 48 Tokyo references used for generation were annotated before any generation (the other 8 afterwards, \S{}3); the London references were annotated, audited and frozen (dataset sha256 in each preregistration) before any London generation, and selected from the qualified pool by a fixed rule --- 15 frames with large non-anchor text and 15 drawn with a fixed seed (\texttt{select\_\allowbreak{}signs.\allowbreak{}py}) for round A, the remaining qualified references for round B.

\begin{table}[!htbp]
\centering
\footnotesize
\setlength{\tabcolsep}{3pt}
\begin{tabularx}{\linewidth}{>{\raggedright\arraybackslash}l>{\raggedright\arraybackslash}X>{\raggedright\arraybackslash}X>{\raggedright\arraybackslash}X}
\toprule
\textbf{Set} & \textbf{References} & \textbf{Quadrilateral drawn by hand / accepted from proposal} & \textbf{Name typed or corrected / OCR pre-fill kept} \\
\midrule
Tokyo & 56 & 19 / 37 & 25 / 31 \\
London round A & 30 & 5 / 25 & 9 / 21 \\
London held-out (floor F1; round B is 40 of these) & 41 & 7 / 34 & 11 / 30 \\
\bottomrule
\end{tabularx}
\end{table}

\textbf{Attribution.} Every reference is derived from a Mapillary image under CC BY-SA 4.0. The 127 references come from 116 images by five contributors (cernbyln, mapconcierge, mura, potaro67v, trolleway). Re-projecting an equirectangular panorama to a perspective view is a modification, and every released record says so: \texttt{release/\allowbreak{}attribution.\allowbreak{}json} carries, per reference, the image ID and page URL, contributor and profile URL, capture time, licence, \texttt{modified: true}, and the re-projection parameters (field of view, yaw, pitch). The per-reference table is \texttt{tables/\allowbreak{}appendix\_\allowbreak{}a\_\allowbreak{}attribution.\allowbreak{}md}. For six London references whose camera was changed during annotation, the frozen dataset kept only the image ID; their contributor and capture time are filled from the capture-time metadata of the same image ID, and the table lists them. The frozen datasets themselves are not edited, so their recorded hashes still hold.

\section*{S9 Prompt template and per-vendor request parameters}\label{sec:S9}\addcontentsline{toc}{section}{S9 Prompt template and per-vendor request parameters}

All requests were built by one prompt builder shared by every vendor (\texttt{bench\_\allowbreak{}generate.\allowbreak{}py}), and are reproduced verbatim from the generation records in \texttt{tables/\allowbreak{}appendix\_\allowbreak{}b\_\allowbreak{}prompts.\allowbreak{}md}. The text is identical across Seedance 2.0, 2.0-mini and Wan 3.0 for every Tokyo shot. Veo's prompts differ only in the timeline, rewritten for its 8-second clip (0--2.4 s wide view, 2.4--5.6 s slow push-in, 5.6--8.0 s locked final frame); the naming-ablation prompts differ only in the removed naming sentence; London prompts differ only in the anchor's name. The negative-list sentence quoted in \S{}5.1 is present in all 380 recorded requests (192 Tokyo, 48 naming ablation, 140 London).

\begin{table}[!htbp]
\centering
\footnotesize
\setlength{\tabcolsep}{3pt}
\begin{tabularx}{\linewidth}{>{\raggedright\arraybackslash}l>{\raggedright\arraybackslash}X>{\raggedright\arraybackslash}X>{\raggedright\arraybackslash}X>{\raggedright\arraybackslash}X>{\raggedright\arraybackslash}X}
\toprule
\textbf{Set} & \textbf{Requests} & \textbf{Model} & \textbf{Conditioning} & \textbf{Duration} & \textbf{Seed} \\
\midrule
Seedance 2.0 & 48 & \texttt{doubao-\allowbreak{}seedance-\allowbreak{}2-\allowbreak{}0-\allowbreak{}260128} & reference image & 5 s & per shot \\
Seedance 2.0-mini & 48 & \texttt{doubao-\allowbreak{}seedance-\allowbreak{}2-\allowbreak{}0-\allowbreak{}mini-\allowbreak{}260615} & reference image & 5 s & per shot (same as 2.0) \\
Wan 3.0 (Tokyo, naming ablation, London A/B) & 48 + 48 + 30 + 40 & \texttt{wan3.\allowbreak{}0-\allowbreak{}video-\allowbreak{}prime} & reference image & 5 s & per shot \\
Veo 3.1 Fast (Tokyo, London A/B) & 48 + 30 + 40 & \texttt{veo-\allowbreak{}3.\allowbreak{}1-\allowbreak{}fast-\allowbreak{}generate-\allowbreak{}preview} & reference image & 8 s & none \\
\bottomrule
\end{tabularx}
\end{table}

Each shot's seed is derived from a hash of its original request, so every vendor that accepts a seed, and the naming ablation, received the same seed for the same shot. Requests that failed at the network level were resubmitted with unchanged settings; where that happened, the first-run record is kept next to the job's \texttt{generation.\allowbreak{}json}.

\section*{S10 Rater instructions and task packs}\label{sec:S10}\addcontentsline{toc}{section}{S10 Rater instructions and task packs}

\textbf{ATF and probes} (\texttt{human\_\allowbreak{}eval.\allowbreak{}py}). Raters see the reference crop and the frame or probe crop rectified to the same canvas, with no model name, automatic score or ground-truth string, in random order. They score 1--5: 5 identical (both readable and the same characters; sharpness and lighting do not matter), 4 slightly degraded, 3 recognisably the same sign with some characters damaged, 2 garbled (still in place but no longer characters), 1 rewritten (reads as a different string: substituted, added, dropped or reordered characters) --- or 0 to abstain when neither side can be read. A second question asks whether the two read as the same string in order: same / different and the frame's text is sharp / different but blurred / the reference itself cannot be read. Raters are told to judge only from the two images, not from knowing the shop's real name.

\textbf{Regions} (\texttt{region\_\allowbreak{}eval.\allowbreak{}py}). Raters see the same location cropped from the reference and from the frame, the region outlined in yellow, and choose: consistent (same text, or both unreadable and the frame no clearer); changed characters; text where the reference has none or cannot be read; pseudo-glyphs (looks like text but is not characters); or cannot judge (misaligned, occluded, or not text). The three middle answers count as fabricated. Sampling is described in \S{}4.4.

\textbf{Position-free search} (\texttt{find\_\allowbreak{}eval.\allowbreak{}py}, S5). Raters see a generated region outlined and the whole reference, and answer whether the same text appears anywhere in the reference: yes / no / the reference is too blurred to tell / the outlined region is not text.

Each task pack (\texttt{tasks.\allowbreak{}json}, released with the crops and ratings) records its sampling seed, the item order, and --- hidden from raters --- each item's set, shot and automatic label.

\section*{S11 Preregistrations}\label{sec:S11}\addcontentsline{toc}{section}{S11 Preregistrations}

For the benchmark experiments, fifteen preregistrations were written and committed before the data they govern existed or were inspected. Each file ends with its result section and any correction, appended afterwards without altering the registered text; the git history is the record of when each part was written. They are released as registered (in Chinese, the working language of the project), except that a name was removed in three places, listed in \texttt{release/\allowbreak{}PREREG-\allowbreak{}REVIEW.\allowbreak{}md}. \texttt{tables/\allowbreak{}appendix\_\allowbreak{}f\_\allowbreak{}prereg.\allowbreak{}md} lists each file with its first and last commit.

\begin{table}[!htbp]
\centering
\footnotesize
\setlength{\tabcolsep}{3pt}
\begin{tabularx}{\linewidth}{>{\raggedright\arraybackslash}l>{\raggedright\arraybackslash}X}
\toprule
\textbf{File} & \textbf{Registers} \\
\midrule
\texttt{probes-\allowbreak{}validation-\allowbreak{}2026-\allowbreak{}10-\allowbreak{}01.\allowbreak{}md} & Probe hold-out round: four predictions (ATF validation) \\
\texttt{mini-\allowbreak{}vs-\allowbreak{}2.\allowbreak{}0-\allowbreak{}2026-\allowbreak{}10-\allowbreak{}01.\allowbreak{}md} & Seedance 2.0 vs 2.0-mini, paired comparison \\
\texttt{prompt-\allowbreak{}ablation-\allowbreak{}smoke-\allowbreak{}2026-\allowbreak{}10-\allowbreak{}02.\allowbreak{}md} & Smoke round removing the sign-fidelity instructions \\
\texttt{peripheral-\allowbreak{}fabrication-\allowbreak{}2026-\allowbreak{}10-\allowbreak{}05.\allowbreak{}md} & Peripheral fabrication, string-level measure (failed; S3) \\
\texttt{peripheral-\allowbreak{}regions-\allowbreak{}2026-\allowbreak{}10-\allowbreak{}06.\allowbreak{}md} & Peripheral fabrication by text regions: automatic blind test and human rating \\
\texttt{peripheral-\allowbreak{}wan-\allowbreak{}heldout-\allowbreak{}2026-\allowbreak{}10-\allowbreak{}07.\allowbreak{}md} & First held-out data: Wan 3.0, all shots \\
\texttt{veo-\allowbreak{}heldout-\allowbreak{}2026-\allowbreak{}10-\allowbreak{}07.\allowbreak{}md} & Veo 3.1 Fast, all shots: second held-out vendor \\
\texttt{unname-\allowbreak{}ablation-\allowbreak{}2026-\allowbreak{}10-\allowbreak{}08.\allowbreak{}md} & Naming ablation \\
\texttt{baseline-\allowbreak{}and-\allowbreak{}geometry-\allowbreak{}2026-\allowbreak{}10-\allowbreak{}09.\allowbreak{}md} & Existing-metric baseline (VBench-I2V) and position checks \\
\texttt{size-\allowbreak{}strata-\allowbreak{}2026-\allowbreak{}10-\allowbreak{}10.\allowbreak{}md} & Is the anchor--periphery gap only text size? \\
\texttt{london-\allowbreak{}crossscript-\allowbreak{}2026-\allowbreak{}10-\allowbreak{}11.\allowbreak{}md} & Latin-script replication, London, round A \\
\texttt{r1-\allowbreak{}multiscale-\allowbreak{}2026-\allowbreak{}10-\allowbreak{}14.\allowbreak{}md} & Region measure R1: held-out floors and Tokyo re-scoring \\
\texttt{london-\allowbreak{}round-\allowbreak{}b-\allowbreak{}2026-\allowbreak{}10-\allowbreak{}15.\allowbreak{}md} & Latin-script replication, London, round B \\
\texttt{r1-\allowbreak{}tolerance-\allowbreak{}2026-\allowbreak{}10-\allowbreak{}15.\allowbreak{}md} & R1 support-tolerance sensitivity \\
\texttt{london-\allowbreak{}human-\allowbreak{}2026-\allowbreak{}10-\allowbreak{}16.\allowbreak{}md} & Human check of the London regions \\
\bottomrule
\end{tabularx}
\end{table}

\clearpage
\section*{S12 Human calibration of readability and text correctness}\label{sec:S12}\addcontentsline{toc}{section}{S12 Human calibration of readability and text correctness}

\textbf{Scope.} This section reports a separate development study, not part of the benchmark: it calibrates OCR-based readability and correctness labels against two human raters on other data, and it changes no number in the main text. It bears on the main study only through \S{}5.3: string- and OCR-based labels mishandle text that humans find unreadable, which is the form most peripheral fabrication takes.

\textbf{Design.} The rating protocol was frozen (commit \texttt{b0f2687} of the project repository) before any new judgment was collected, on six reference images and 108 existing clips from Wan 2.2 (5B) and Wan 3.0. Two raters each completed six sessions and made 890 judgments, including repeats. Each rater judged 341 distinct input crops and 467 distinct output crops. We sampled outputs at 1.5 seconds, stratified by automatic label, model and degradation level. We kept all outputs automatically labelled as wrong, sampled up to 25 from each other stratum, and used inverse inclusion probabilities to estimate precision and recall in the development pool. Repeats were reserved for assessing repeatability. The interface hid the reference transcriptions, model identities, degradation levels, seeds and automatic labels, and sessions ran from the heaviest degradation to the lightest; the same rater could still see several versions of a region across sessions, so this is a development calibration under interface hiding, not a fully blind study. The first rater was the author and had seen the pilot results and examples with the unblurred references before the study. That rater completed the first 21 items before the cropping-error instructions were clarified. The second rater is the second rater of the main study (\S{}7). The protocol, crops and ratings are not part of the benchmark release.

\textbf{Second rater's declaration --- a protocol deviation.} The protocol required the second rater to declare, before rating began, any prior exposure to the materials. The written declaration was obtained only after scoring had ended: it is dated 23 November 2026, and scoring ended on 18 November. In it the second rater reported no prior exposure to the six reference images, the ground-truth transcriptions or the pilot results, reported fluency in reading English, and stated an intention to score and submit independently. We report it as written and as a deviation.

\textbf{Results.} The strict rule used exact transcription matching after normalizing character forms, whitespace, case and selected punctuation. Agreement on the resulting output labels was $\kappa = 0.71$. Using the sampling weights, the automatic rule recovered only 24.1\% of outputs the first rater judged unreadable and 24.2\% of those the second rater judged unreadable, although when it did call an output unreadable it was usually right (precision 96\% and 93\%). Its ``correct'' label was over-assigned: only 53\% and 60\% of outputs it called correct were judged correct by the first and second rater. Recall for outputs judged readable but incorrect was 5.0\% and 4.6\%, respectively. Under the legacy rule, which also allows character variants and substring matches, both recall estimates were 100\%, with four human-positive outputs for the first rater and five for the second; the outcome for this class is therefore decided mainly by the matching rule.

\textbf{Clear but incorrect text.} This check required an unreadable input crop and a readable output with a transcription that disagreed with the reference. Of the 28 outputs flagged automatically, 16 had a determinate joint human label, and the raters jointly confirmed three, all from a single reference image: a precision of 3/16 = 18.8\%, or 10.7--53.6\% if every undetermined output is counted as confirmed or as rejected. The weighted prevalence of the event was 0.2\% and 0.4\% for the two raters. These results characterize the automatic rule on the six-image development set at the sampled time point and do not show that the event is absent elsewhere.
\end{document}